\documentclass[12pt]{article}

\usepackage{amssymb}
\usepackage{amsmath}

\usepackage{natbib}
\usepackage{stfloats}
\usepackage{amsmath,amssymb}
\usepackage{mathtools}
\usepackage{xspace}
\usepackage{booktabs}
\usepackage{xcolor}
\usepackage{colortbl}
\usepackage{multirow}
\usepackage{array}
\usepackage{url}
\usepackage{makecell}
\usepackage{changepage}
\usepackage{changes}

\usepackage{framed}
\usepackage{float}
\usepackage{booktabs}
\usepackage{longtable}
\usepackage{nicefrac}
\usepackage{amssymb}
\usepackage{amsmath}
\usepackage{xspace}
\usepackage{mathtools}
\usepackage{diagbox}
\usepackage{graphics}
\usepackage{rotating}
\usepackage[algo2e,ruled,vlined,linesnumbered]{algorithm2e}

\usepackage{subfigure}
\usepackage{ifthen}
\usepackage{wrapfig}
\usepackage{balance} 
\usepackage{epsfig}
\usepackage{xcolor, colortbl}
\usepackage[normalem]{ulem}
\usepackage{dsfont}
\usepackage{graphicx}
\usepackage{hyperref}

\usepackage{authblk}
 \hypersetup{
     colorlinks=true,
     linkcolor=blue,
     filecolor=blue,
     citecolor = blue,      
     urlcolor=blue,
     }

 \usepackage{comment}
 \usepackage{pdflscape}
 \usepackage{afterpage}
 \usepackage{booktabs, multirow} 
 \usepackage{soul}

\title{A Survey of Features Used for Representing Black-box Single-objective Continuous Optimization}

\author[1,2]{Gjorgjina Cenikj}
\author[1,2]{Ana Nikolikj}
\author[1,2]{Ga\v{s}per Petelin}
\author[3]{Niki van Stein}
\author[4]{Carola Doerr}
\author[1]{Tome Eftimov}

\affil[1]{Computer Systems Department, Jožef Stefan Institute, Ljubljana, Slovenia}
\affil[2]{Jožef Stefan International Postgraduate School, Ljubljana, Slovenia}
\affil[3]{Leiden Institute for Advanced Computer Science, Leiden, The Netherlands}
\affil[4]{Computer Science department LIP6, Sorbonne Université, CNRS, Paris, France}

     \date{January 2026}   
\begin{document}
\maketitle
\begin{abstract}
This survey examines key advancements in designing features to represent optimization problem instances, algorithm instances, and their interactions within the context of single-objective continuous black-box optimization. These features support machine learning tasks such as algorithm selection, algorithm configuration, and problem classification, and they are also used to evaluate the complementarity of benchmark problem sets. We provide a comprehensive overview of problem landscape features, algorithm features, high-level problem-algorithm interaction features, and trajectory features, including the latest works from the past five years. We also point out limitations of the current state-of-the-art and suggest directions for future research. 

\end{abstract}





\section{Introduction}

It is widely acknowledged that different instances of an optimization problem and different performance criteria require different algorithm instances for optimal resolution. This is particularly evident in computationally difficult problems, where the choice of the algorithm used to solve it has a major impact on the quality of the final solution or the time needed to find a satisfactory one. Harnessing this performance complementarity has been the target of several approaches, with one of the most prominent ones being \textit{per-instance algorithm selection}~\cite{as_survey}, which aims at mapping problem instances to the most suitable among a portfolio of different algorithm instances.

Per-instance algorithm selection is a challenging problem by itself, which has attracted significant research interest in recent years~\cite{as_survey,as_combinatorics, munoz2015algorithm}.  Most studies in this area use meta-learning~\cite{vanschoren2019meta} to train a supervised Machine Learning (ML) model to predict the performance of an algorithm instance or to directly predict the algorithm which is expected to work best on the input problem. This process requires input features for the ML model that are linked to the algorithm instance. The features are typically related to the problem instance landscape~\cite{gallagher2024towards}. 

Other learning tasks that leverage problem landscape features include problem classification and analyzing the complementarity of benchmark problem suites. In problem classification, an ML model learns to identify \textit{which} problem is represented or \textit{its difficulty} based on landscape features~\cite{naudts2000comparison,renau2019expressiveness,he2007note}. A related task is to classify key problem properties like ruggedness, neutrality, and gradients. For complementarity analysis~\cite{skvorc2022transfer}, unsupervised learning techniques such as clustering and dimensionality reduction are applied to evaluate the coverage of benchmark suites, which are then used to select representative learning data necessary for statistical benchmarking or developing robust ML models~\cite{selector,lang2021exploratory,vermetten2025standardized}.

However, recently, features related to the properties/characteristics of the algorithm instances or features extracted from the interaction between an algorithm instance and a problem instance (i.e., high-level interaction features and trajectory-based features) have also been used and incorporated into various ML applications~\cite{10.1145/3583131.3590401, opt2vec, algorithm_features_time_series_cmaes, ochoa2014local}. 



In this survey, we provide an overview of features used to represent problem landscapes, algorithm instances, and their interactions in the field of single-objective continuous optimization in a high-level interaction level and a trajectory-based level. We explore their applications in tasks such as algorithm selection, problem classification, and evaluating the complementarity of benchmark problem suites. Additionally, we identify research gaps and suggest directions for future development.

Throughout this survey, we use the following notation to maintain consistency across mathematical expressions. Let $\mathbf{x} = (x_1, x_2, \ldots, x_d)$ denote a candidate solution vector in a $d$-dimensional decision space $\mathcal{X} \subseteq \mathbb{R}^d$, and let $f(\mathbf{x})$ represent the objective function value associated with $\mathbf{x}$. A set of $n$ candidate solutions sampled from the search space is denoted by $\{\mathbf{x}^{(1)}, \mathbf{x}^{(2)}, \ldots, \mathbf{x}^{(n)}\}$, with corresponding objective values $f(\mathbf{x}^{(1)}), f(\mathbf{x}^{(2)}), \ldots, f(\mathbf{x}^{(n)})$. The trajectory of an optimization algorithm executed for a budget of $b$ iterations (or $t$ function evaluations), consists of all the candidate solutions explored by the algorithm $\{\mathbf{x}^{(1)}, \mathbf{x}^{(2)}, \ldots, \mathbf{x}^{(t)}\}$, with corresponding objective values $f(\mathbf{x}^{(1)}), f(\mathbf{x}^{(2)}), \ldots, f(\mathbf{x}^{(t)})$.

\subsection{Relation To Other Surveys}
While our focus is on single-objective continuous optimization, we acknowledge that a few studies have begun extending these concepts to multi-objective optimization~\cite{ochoa2024funnels,liefooghe2023pareto}, constrained multi-objective optimization~\cite{liefooghe2019landscape,alsouly2024online}, and mixed-integer optimization~\cite{prager2023investigating,dietrich2024hybridizing}, as well as surveys focused on the automated design of algorithms~\cite{10993463}. However, this survey does not cover those areas, leaving room for specialized communities to present their own reviews. The methodologies for calculating and learning meta-features discussed here can also be easily adapted to inspire research in other optimization fields that currently lack extensive studies.

We also acknowledge that there are other survey papers addressing the same topic~\cite{malan2013survey,malan2021survey} and tutorials~\cite{liefooghe2023pareto,kerschke2023exploratory}.
However, in this survey, we chose not to include an extensive list of references from these works (readers should check these works for a more comprehensive view). Instead, our search process and inclusion criteria were the following. We focused on recent papers and trends from the last five years, sourced from leading optimization conferences and their special sessions, such as the Genetic and Evolutionary Computation Conference (GECCO), IEEE Congress on Evolutionary Computation (IEEE CEC), Parallel Problem Solving from Nature (PPSN), and EvoStar. Additionally, we reviewed relevant journal articles from this period and examined contributions from major ML conferences, including the International Conference on Learning Representations (ICLR), International Joint Conferences on Artificial Intelligence (IJCAI), and the Conference on Neural Information Processing Systems (NeurIPS).

We included works that used optimization meta-features for at least one of the following: algorithm selection (classification, regression, or ranking), performance prediction (per-algorithm or per-instance), problem classification (problem class, difficulty, or high-level landscape properties), and complementarity or visualization of benchmark suites (for example, clustering, dimensionality reduction).
We considered studies reporting results on established suites and modern generators, where feature learning or evaluation was central. We excluded works outside the primary scope (purely multi-objective, constrained multi-objective, or mixed-integer) unless they directly informed single-objective feature methods, papers without extractable experimental details (for example, no identifiable benchmarks or metrics), and optimization studies which do not include feature representations of optimization problems, algorithms, or trajectories.

Furthermore, works such as~\cite{malan2013survey,malan2021survey} provided valuable taxonomies but offered limited attention to algorithm features and completely omitted trajectory-based approaches, since these were not developed at the time. In contrast, our survey introduces a structured taxonomy that integrates four representation families: problem landscape features, algorithm features, high-level interaction features, and trajectory-based features, thereby broadening the scope substantially.
On the other hand, tutorial contributions such as~\cite{liefooghe2023pareto,kerschke2023exploratory} played an important pedagogical role by explaining the basic concepts of landscape analysis and providing practical guidance on tools (for example, software for computing ELA features). However, these tutorials were not intended to provide a comprehensive synthesis of the state of the art, and they do not cover the surge of recent methods, particularly those based on deep learning (for example, CNN-based fitness maps, TransOpt, Deep-ELA, DoE2Vec) and trajectory-based representations (for example, DynamoRep, Opt2Vec, probing trajectories).
Finally, unlike these earlier contributions, our survey emphasizes critical evaluation and open challenges. We analyze methodological issues such as over-reliance on the BBOB benchmark, lack of variance and effect size reporting, weak cross-benchmark generalization, and limited interpretability of black-box features - points that earlier surveys and tutorials did not address. We also outline forward-looking directions including invariance-aware feature design, and cross-benchmark evaluation protocols.
In summary, our survey advances the field by unifying problem, algorithm, interaction, and trajectory representations, incorporating modern deep learning approaches, and highlighting methodological rigor and generalization, offering a comprehensive and up-to-date overview.

\section{Machine Learning Pipeline using Optimization Features}

\begin{figure*}
    \centering
    \includegraphics[width=\textwidth]{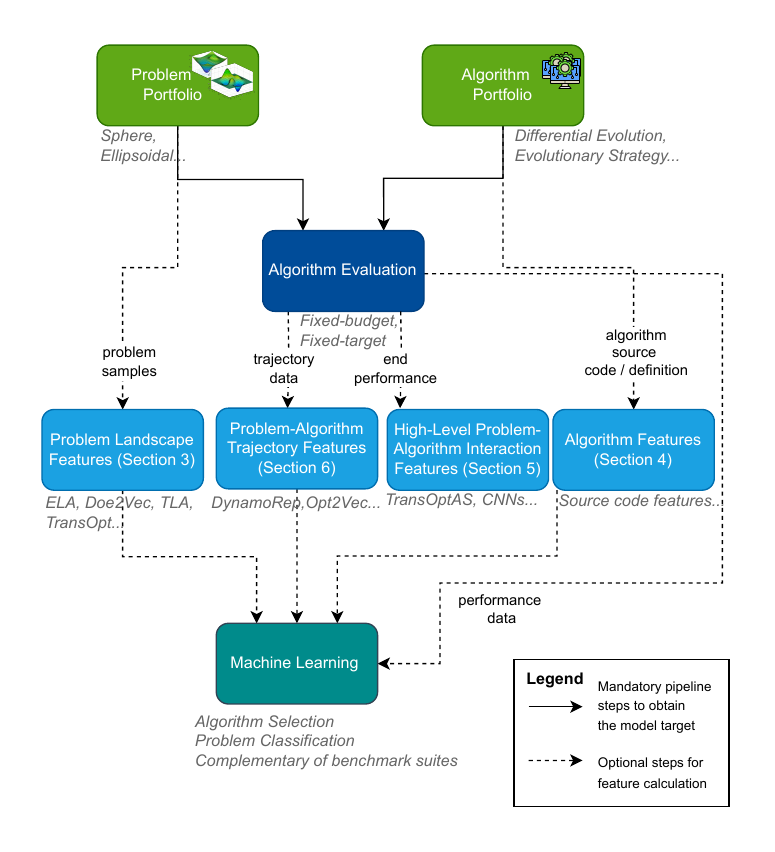}
    \caption{A typical machine learning pipeline using optimization meta-features}
    \label{fig:algorithm_selection_pipeline}
\end{figure*}

Figure~\ref{fig:algorithm_selection_pipeline} illustrates an ML pipeline that includes essential components for calculating or learning features related to problem landscapes, algorithm characteristics, and problem-algorithm interactions. It also highlights several learning tasks - such as algorithm selection, problem classification, and complementarity analysis of benchmark problem suites - where these features serve as inputs to the ML model. It is important to note that the specific features and components involved may vary depending on the particular ML task being solved.

The main components of such a pipeline are the following:
\begin{itemize}
    \item \textbf{Problem Portfolio} - A set of optimization problem instances for which we want to find an optimal algorithm instance, analyze their coverage, or predict their interoperable properties. In the single-objective optimization domain, examples of such functions are the sphere function, ellipsoidal functions, discus, etc. These problem instances are structured in problem benchmark suites, the most commonly used ones being the Black Box Optimization Benchmarking (BBOB)~\cite{bbob} suite and the IEEE Congress on Evolutionary Computation (CEC) Special Sessions and Competitions on Real-Parameter Single-Objective Optimization~\cite{cec2013, cec2014, cec2015, cec2017} suites. For a sound (academic) evaluation of the ML models, the problem portfolio should cover problem instances that are diverse, challenging, and discriminating of algorithms’ performances. In recent years, there has been a growing interest in the optimization field regarding the generation of new optimization problem instances in an automatic manner~\cite{munoz2020generating,tian2020recommender,gaviano2003software,prager2023neural, dietrich2022increasing,MABBOBteloversion}. Generating novel problem instances has the potential to introduce previously unexplored optimization challenges in ongoing research. 

    
    \item \textbf{Algorithm Portfolio} - A set of optimization algorithm instances that are candidates for solving an optimization problem instance. Some examples in single-objective black-box optimization are Differential Evolution~\cite{StornPriceDE}, Covariance Matrix Adaptation Evolution Strategy (CMA-ES)~\cite{HansenO01}, and Particle Swarm Optimization (PSO)~\cite{pso_algorithm} algorithms. In addition, modular optimization frameworks are also utilized such as ModCMA~\cite{ModCMAnew,van2016evolving}, ModDE~\cite{vermetten2023modular}, and PSO-X~\cite{camacho2021pso}.  
    
    \item \textbf{Algorithm Evaluation} - The process of running the optimization algorithm instances on the optimization problem instances and recording their performance in terms of some evaluation/performance metric. Two standard evaluation scenarios include the fixed-budget and fixed-target evaluations. In the \textbf{fixed-budget scenario,} the algorithm instance is allowed to use a predefined budget and one studies the solution quality or, if the value of the global optimum is known, the \textit{target precision,} defined as the difference between the quality of the solution recommended by the algorithm (typically, but not necessarily, the best evaluated one) and that of the global optimum. In the \textbf{fixed-target scenario,} the algorithm is tasked with finding a solution of a certain quality, and we study the budget that is needed by the algorithms to find a solution of at least this quality.  We recall that, in black-box optimization, the \textit{budget} is often specified in terms of function evaluations or iterations, not in terms of (CPU or wall-clock) time. A third commonly used set of performance measure considers the \textit{anytime performance} of the algorithms, measured through empirical cumulative distribution functions~\cite{COCOanytime} or directly through empirical attainment functions~\cite{EAFIOH}.   

    \item \textbf{Problem Landscape Features} - The description of an optimization problem instance in terms of numerical features. Landscape analysis is a line of research whose aim is to extract meta-features that describe the properties of the problem instances~\cite{zou2022survey}. We typically distinguish between two levels of landscape features: high-level properties of the fitness landscape that are qualitative descriptors of the landscape structure (e.g., multi-modality, separability, search space homogeneity)~\cite{malan2013survey} and low-level features  (i.e., exploratory landscape analysis, ELA~\cite{ela} such as candidate solution distribution, local search, convexity, meta-model, smoothness, ruggedness) calculated using different statistical methods from a sample of candidate solutions sampled from the decision space for a given problem instance using some sampling technique). We need to highlight here that those features are calculated to describe the whole landscape space of a problem instance, without taking care about the landscape visited/observed when an algorithm instance is run on it. Problem landscape features therefore describe intrinsic properties of optimization problems, independently of any algorithmic influence.
    We provide an overview of problem landscape feature construction in Section~\ref{sec:problem_landscape_features}.

    \item \textbf{Algorithm Features} - These features characterize the algorithm instance, independently of the problem it is used to optimize. They are generated by analyzing the source code~\cite{algorithm_code_features} or derived from algorithm parameters that remain constant and do not depend on the problem being solved. It is the low resourced investigated feature group.  An overview of algorithm features is given in Section~\ref{sec:algorithm_representation}.
    
    \item \textbf{High-level Problem-Algorithm Interaction Features} - These features rely on high-level information about the optimization problems on which the algorithms are executed, rather than capturing the trajectory of the optimization runs. In essence, these features connect high-level information about a set of optimization problems with the final performance achieved by an algorithm. Unlike algorithm features, which describe only the algorithm, without any influence of the problem on which it is executed, the high-level problem-algorithm interaction features capture properties of both the problem and the algorithm. 
    
    Different approaches exist focusing on features based on their performance~\cite{performance2vec}, features based on graph embeddings~\cite{anak_kg_performance_prediction} and graph neural networks (GNNs)~\cite{kostovska2025geometric}, and features based on configuration settings for modular algorithm frameworks~\cite{nikolikj2024quantifying,van2025explainable}. Section~\ref{sec:high-level-problem-algorithm_representation} provides an overview of the high-level problem-algorithm features.

    \item \textbf{Trajectory-based Features for Capturing Problem-Algorithm Interaction} - Problem landscape features characterize problem instances in terms of numerical features calculated on a sample from the entire problem instance. This step is typically done \textit{before} the optimization algorithm is run, meaning that additional function evaluations are required for the feature construction process. Additionally, the high-level problem-algorithm interaction features encode information about the problem landscape and the algorithm's final performance. 
    An alternative line of work is the usage of samples generated by optimization algorithms during their execution on a problem instance. This type of representation would capture interactions between the problem instance and the optimization algorithm. An overview of trajectory-based feature constructions for capturing problem-algorithm interaction is given in Section~\ref{sec:trajectory_based_representation}. 
    The difference between trajectory-based features and high-level problem-algorithm interaction features is in the input data used for their calculation. Both feature types are dependent on both the problem and the algorithm. However, the trajectory features are calculated using the entire trajectory of candidate solutions explored during algorithm execution. They therefore capture the search behaviour of the algorithm executed on a problem. On the other hand, the high-level problem-algorithm interaction features do not use the entire trajectory as input, but instead may rely only on the final performance of the algorithm. These features therefore do not capture the search behaviour, but the relation between the characteristics of a problem, and the performance of an algorithm.
    
    \item \textbf{Machine Learning} - The problem landscape features, algorithm features, high-level problem-algorithm interaction features, and trajectory-based features, or a combination of different groups can be used as input data for an ML model in different learning tasks. The three most researched tasks are presented below:
    \begin{itemize}
        \item Algorithm Selection -- One approach for training an AS model is to treat it as a \textit{multi-class classification problem}~\cite{skvorc2022transfer}, with the objective of identifying a single optimal algorithm for the specific problem instances. Alternatively, it can be treated as a \textit{multi-label classification problem}~\cite{mlc}, where multiple algorithms may be selected (e.g., two or three different algorithms with similar performance). Other learning scenarios include solving a \textit{regression problem}~\cite{anan_ela_feature_importance, anak_ela_feature_importance, risto_ela_feature_importance, anan_ela_feature_importance_2}, in which the ML model predicts the performance achieved by each algorithm, or \textit{ranking}~\cite{gasper_affine_generalization}, where the aim is to rank the algorithms according to a specific evaluation metric. A comprehensive study comparing approaches for algorithm selection in single-objective optimization -- through different learning tasks (multi-class classification, pairwise classification, regression) and performed with different ML algorithms (i.e., Random Forest~\cite{biau2016random}, XGBoost~\cite{chen2016xgboost}, TabPFN~\cite{hollmann2022tabpfn}, and FT-Transformer~\cite{gorishniy2021revisiting}) -- is available in~\cite{10.1145/3583133.3590697}.   
        \item{Problem Classification} -- In this task, meta-features serve as input to a model to predict either the problem represented (for problem landscape features) or the problem being solved (for problem-algorithm trajectory features)~\cite{10.1145/3583131.3590401,korovsec4316939opt2vec}. Identifying this information can assist users and researchers in applying their domain knowledge to enhance the optimization process.
        \item {Complementarity of benchmark problem suites} -- In this task, various sets of problem instances are represented by problem landscape features, and their diversity is examined using unsupervised learning techniques like clustering~\cite{dietrich2024impact}, dimensionality reduction~\cite{urban_complementary_analysis}, and graph algorithms~\cite{selector}.
        \end{itemize}

\end{itemize}

It is important to note that in this survey, we will focus on various methodologies used to calculate or learn problem landscape features, algorithm features, and problem-algorithm trajectory features, along with their existing applications. However, other components, such as the selection of problem and algorithm portfolios, and different approaches to ML modeling across various learning tasks, are beyond the scope of this survey.

\section{Problem Landscape Features}
\label{sec:problem_landscape_features}

\begin{figure}
    \centering
    \includegraphics[width=\textwidth]{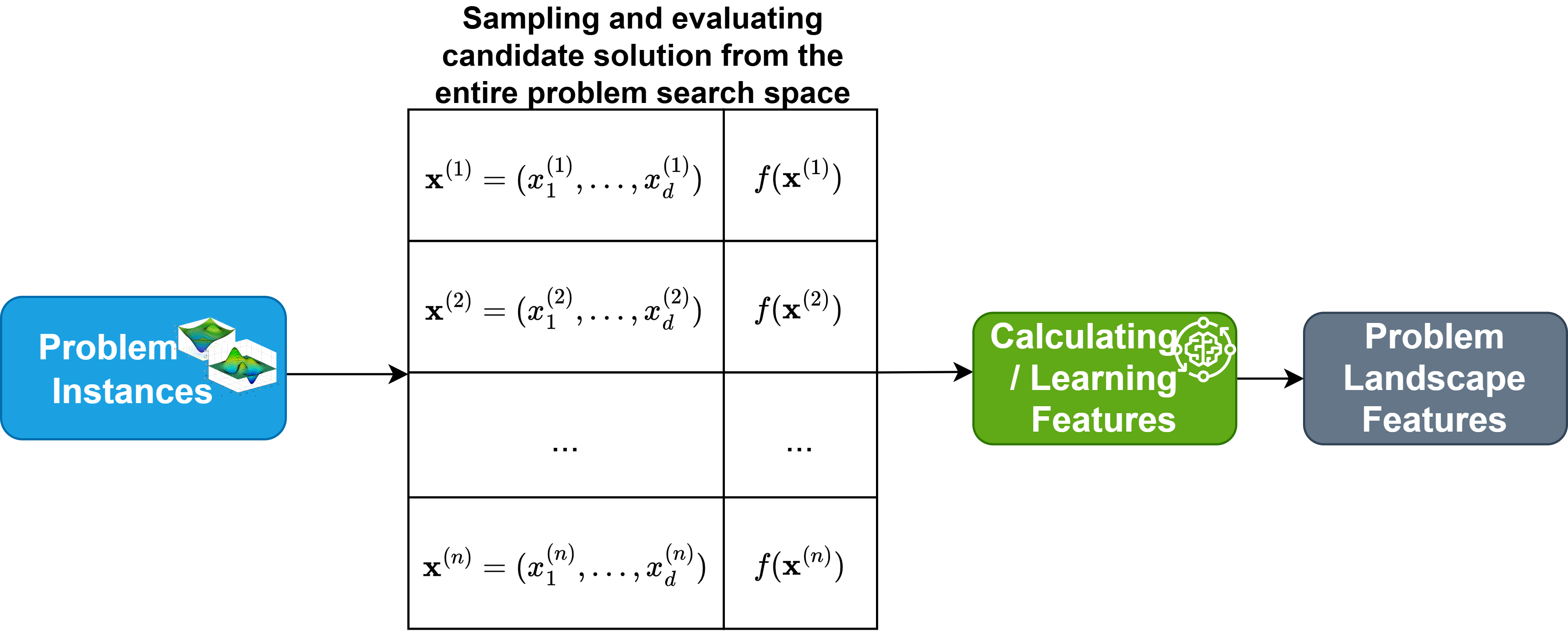}
    \caption{The general pipeline of calculating problem landscape features.}
    \label{fig:pip_problem}
\end{figure}

In this section, we describe approaches that calculate problem landscape features using a set of candidate solutions, sampled from the decision space of the problem instance through non-adaptive or deterministic sampling techniques. These features are characterized by their aim to represent the problem instance, regardless of which algorithm is executed.
Figure~\ref{fig:pip_problem} illustrates the general pipeline of calculating problem landscape features.

Several approaches for capturing characteristics/properties of single-objective continuous optimization problems have already been proposed. They can be broadly categorized as providing high-level features that are more interpretable and low-level features such as in the case of \textit{Exploratory Landscape Analysis}~\cite{ela}, \textit{Topological Landscape Analysis}~\cite{tla}, and \textit{deep learning-based approaches}~\cite{seiler2022collection, transoptas}. The latter can be learned through supervised learning (which includes algorithm performance data) or unsupervised learning (which uses only samples of the problem landscape space obtained through a sampling technique). 

\subsection{High-level landscape features}
High-level fitness landscape analysis captures optimization problem characteristics such as the degree of global structure (none, low, medium, high), separability, variable scaling, multimodality, and search space homogeneity. For example, features related to ruggedness, neutrality, gradients, global landscape structure, deception, and searchability have been explored together with their correlations with the diversity rate of change metrics of the behavior of the particle swarm algorithm~\cite{engelbrecht2021influence}. The results have shown links between particular features and PSO behavior. In contrast to some of the low-level landscape features, these features are designed to be highly interpretable.  

\subsection{Low-level landscape features}

\subsubsection{Exploratory Landscape Analysis}
Exploratory landscape analysis (ELA)~\cite{ela} is an approach to characterize black-box optimization problem instances using numerical measures that each describe a different aspect of the problem instances. 
ELA features can be further distinguished into \textit{cheap} features that are computed from a fixed set of samples and \textit{expensive} features that require additional sampling (e.g., they may require running some local search variants or other sequential sampling approaches)~\cite{per_instance_algorithm_configuration}. 
A convenient way to compute ELA features is provided by the \textit{flacco} R package~\cite{flacco}. This package offers 343 different feature values split into 17 feature groups, including \textit{dispersion}, \textit{information content}, \textit{meta-model}, \textit{nearest better clustering}, and \textit{principal component analysis}. Recently, a Python version of the package has been published, offering some additional set of features from local optima networks~\cite{prager2022pflacco}.

The original ELA features~\cite{ela} were grouped into six categories:

\begin{itemize}
    \item The \textbf{convexity} of an optimization problem is captured by observing the difference in the objective value of a point which is a linear combination of two randomly sampled points and the convex combination of the objective values of the two sampled points.

    \item The \textbf{distribution of the objective function values} is characterized by its skewness, kurtosis, and the number of peaks based on a kernel density estimation of the initial design’s objective values.
    
    \item \textbf{Local search} features are extracted by running a local search algorithm and hierarchically clustering the considered solutions in order to approximate problem properties. For instance, the number of clusters is an indicator of multi-modality, while the cluster sizes are related to the basin sizes around the local optima. 

    \item The \textbf{levelset} features split samples into two classes based on whether the value of the objective function falls above or below a certain threshold. Linear, quadratic, and mixture discriminant analysis is used to predict whether the objective values fall below or exceed the calculated threshold. The intuition behind this is that multi-modal functions should result in several unconnected sublevel sets for the quantile of lower values, which can only be modeled by the mixture discriminant analysis method. The extracted low-level features are based on the distribution of the resulting misclassification errors of each classifier. 

    \item The construction of \textbf{meta-model} features requires the use of linear and quadratic regression models, with and without interaction effects, which are fitted to the sampled data. 
    The features are then calculated by taking the adjusted coefficient of determination R2 of the models, indicating model accuracy, as well as the model coefficients. Such features are useful for extracting function characteristics, since functions with variable scaling will not allow a good fit of regression models without interaction effects, and simple unimodal functions could be approximated by a quadratic model.

    \item \textbf{Curvature} features estimate the gradient and Hessians from  samples of the function and use their magnitudes to quantify the curvature.
\end{itemize}

Cell mapping features~\cite{ela_cell_mapping} partition the problem landscape into a finite number of cells:
\begin{itemize}
\item \textbf{Angle} features are based on the location of the worst and best element within each cell since opposite locations of these elements indicate a trend within a cell. More precisely, the features observe the mean and standard deviation of distances from the cell center to the best / worst element, and of the angle between these three points.

\item \textbf{Convexity} features represent each cell by the element closest to the cell center and compare the objective function values of neighboring cells to capture convexity/concavity.

\item \textbf{Gradient homogeneity} features compute the length of the sum of the gradients between each element and its nearest neighbor within the cell. 

\item \textbf{General cell mapping} features interpret the problem landscape cells as absorbing Markov chains. Cells that are visited infinitely often, once they have been visited once, are referred to as periodic cells, while all other cells are transient cells. Different types of features are then extracted based on the probability of moving from one cell to another, the number of cells of different types that appear, the size of the basins of attraction, etc.

\end{itemize}

 Over time, additional ELA features have also been introduced:

\begin{itemize}
    \item \textbf{Basic} features include minimum and maximum of the objective function, number of samples, etc. based on an initial sample of points.
    \item \textbf{Dispersion} features~\cite{ela_dispersion} compare the dispersion among all the samples and among a subset of these points.
    \item \textbf{Linear Model} features~\cite{flacco} first build several linear models with the objective variable being the dependent variable and all the remaining variables being treated as explaining variables. Based on the obtained model coefficients, statistics are calculated that are used as features.
    \item \textbf{Nearest Better Clustering} features~\cite{ela_nbc} are computed by comparing samples to their closest neighbor and to their better closest neighbor.
    \item Features based on \textbf{Principal Component Analysis}~\cite{flacco} describe the relative amount of principal components required to explain a predetermined amount of variability.
    \item \textbf{Barrier Tree} features~\cite{ela_bt} describe the landscape of an optimization function using trees where saddle points are represented with regular nodes while minimums are described by leaves. Features are then generated by computing tree descriptors such as the number of levels and the distance between the leaves.
    \item \textbf{Information Content} features~\cite{ela_ic} try to capture information content of the landscape such as smoothness, and ruggedness. This is achieved by taking into account the distance between neighbours and changes in fitness values between them.
\end{itemize}

The success of ELA can largely be attributed to its accessibility, particularly through the ready-to-use software packages, flacco and pflacco. However, it is not a comprehensive implementation of all existing features. For instance, information epistasis~\cite{seo2007information} or auto-correlation~\cite{weinberger1990correlated}, initially designed for discrete variables, can be easily adapted to continuous ones. In addition, the same is true for the use of entropy~\cite{marin2012landscape} as a measure. 

ELA features are the most commonly used problem landscape features, however they come with a lot of limitations: their poor robustness to sampling strategy and sample size~\cite{lang2021exploratory,renau2019expressiveness,renau2020exploratory}, and invariance to transformation in the problem space (i.e., shifting, scaling, or rotation of a problem instance)~\cite{urban_complementary_analysis,urban_ela_not_invariant,urban_ela_not_invariant_sampling}. Recently, a new study highlights how many of these features are sensitive to absolute objective values (i.e., lack of shift and scale invariance), causing bias in automated algorithm selection models and hindering their ability to generalize to new problem instances~\cite{prager2023nullifying}. To mitigate this, it has been proposed to normalize the sampled data before computing ELA features.

A potential avenue to address the sensitivity of ELA features to sampling strategy and sample size is the exploration of adaptive sampling and aggregation schemes. Instead of fixing the number of samples in advance, one could investigate starting from a small initial design (e.g., Latin hypercube or Sobol) and gradually expanding it in regions where feature estimates appear unstable. Similarly, aggregating features across multiple resamples, possibly with normalization and variance-aware weighting, may offer more stable descriptors.

ELA features have also been used to characterize Neural Architecture Search (NAS)~\cite{elsken2019neural} landscapes across three typical image classification datasets: MNIST, Fashion, and CIFAR-10~\cite{van2020neural}. This analysis reveals distinct characteristics of the NAS landscapes of these three datasets. 
ELA features have also been used to perform complementarity analysis of the hyperparameters optimization landscapes (HPO) and classical BBOB problems from~\cite{bbob}, for black-box optimization~\cite{DoerrDK19} and for XGBoost learner~\cite{schneider2022hpo}. Additionally, the ELA features have been used to identify cheap representative functions of 20 automotive crashworthiness problems which are expensive to evaluate~\cite{cheap_representative_functions_automotive}, where it was also shown that these problems exhibit landscape features different from the classical BBOB benchmark suite.

Overall, ELA is one of the most widely used and tested approaches for characterizing black-box single-objective optimization landscapes, offering a broad set of numerical descriptors. The features are accessible in ready-to-use implementations such as \textit{flacco} and \textit{pflacco}. At the same time, ELA has important limitations: many features are sensitive to the sampling strategy and sample size, and several are not invariant under common transformations of the search or objective space (e.g., shifting, scaling, or rotation).

\subsubsection{Topological Landscape Analysis}
Topological Landscape Analysis~\cite{tla} features characterize optimization problem instances based on Topological Data Analysis~\cite{tda}, which is an approach to analyzing and obtaining features from a finite set of data points, also referred to as a \textit{point cloud}. Similarly to the ELA features, the point cloud is constructed based on a set of samples from the optimization problem. 
Once the samples of the objective function are collected, they are sorted by objective value and divided into multiple layers. The point cloud from each level reveals the structural characteristics specific to that layer. For each point cloud, a persistence diagram is generated, capturing the topological features within the layer. These persistence diagrams are then transformed into finite-dimensional vector representations known as \textit{persistence images}. Persistence images enable each layer’s samples to be represented as a feature vector, where pixel values from the image provide a meaningful encapsulation of the underlying structure in vector form.

This approach is especially robust to noise and supports different metrics for describing the similarity between data points. The goal of Topological Data Analysis is to discover topological structures such as spheres, torus, connected components, or even more complicated surfaces and manifolds, which persist across scales, independently of certain transformations (e.g. rotations, scaling). Discovering such structures can help characterize the point cloud using features that capture the existence of different structures across different scales.

Apart from their original version~\cite{tla}, an improved version called TinyTLA requiring smaller sample sizes for feature calculation has also been proposed in~\cite{tinytla}. 
Here, instead of splitting the problem samples into layers, the information from the entire set of problem samples is captured using a matrix of distances between the candidate solutions ($\mathbf{x}$ values) and objective function ($f(\mathbf{x})$) values of each pair of samples.
The thereby obtained distance matrix is used to further compute the persistence diagrams, persistence images, and feature representations as in the original TLA.
The TinyTLA features have also been tested on the BBOB problem classification task with an extensive analysis of parameter influence and an initial evaluation for the algorithm selection task on the BBOB benchmark.

\subsubsection{Deep Learning-based approaches} Here, more details about low-level features learned by different types of deep neural networks (DNNs) will be provided.

\paragraph{Features learned by using convolutional neural networks} A set of low-level features learned by convolutional neural networks (CNNs)~\cite{gu2018recent} are presented in~\cite{prager2021towards,seiler2022collection}. To calculate them, first, for each problem instance involved in the training data (i.e., BBOB problem instances), a set of candidate solutions is generated using a Latin Hypercube sampling (other sampling techniques can also be used). The set of candidate solutions is further used to calculate a fitness map, which is a two dimensional image with a single channel in a [0,1] range. To generate the image, the objective solution values are normalized in the range of [0,1], and each candidate solution is mapped into a Cartesian plane at the location defined by the decision variables and the objective solution value defines the color range (gray-scaled) of each pixel. One weakness of this transformation is the information loss since two different candidate solutions that are close will be mapped to the same pixel. Transforming the set of candidate solutions into a two-dimensional image allows utilizing a CNN to compute low-level features for the problem instances. In~\cite{prager2021towards}, the CNN that has been utilized is the ShuffleNet v2~\cite{ma2018shufflenet} due to the competitive performance achieved in object classification in computer vision.  In~\cite{seiler2022collection}, the work of using the fitness map and CNN has been extended for high-dimensional data rather than two or three dimensions. For this purpose, four dimensionality reduction techniques have been investigated to reduce the fitness map dimensionality into two dimensions that are further utilized by the CNN to learn the low-level features. Here, 24 BBOB problem classes are used with 150 problem instances per class. The training has been performed in a supervised manner in a problem classification task where the problem instances are classified into one of the three classes based on the high-level properties i.e., multimodality, global structure, and funnel structure. 

\paragraph{Features learned using a point cloud transformer} When calculating the fitness map for high-dimensional problem instances ($d>2$) there is a loss of information from high-dimensional to low-dimensional space. To mitigate this, point cloud transformers have also been utilized to calculate low-level features. In~\cite{seiler2022collection}, the authors modified the original work on point cloud transformers that use the idea of convolutions that operate on the edges within a $kNN$-graph~\cite{wang2019dynamic}. Here, instead of convolutions on the edges, the embedding operates on the node of the $kNN$-graph. This means that every candidate solution is embedded into its local neighborhood (similar candidate solutions). This has been performed since these features have been learned in a supervised manner in a problem classification task where the problem instances are classified into one of the three classes based on the high-level properties i.e., multimodality, global structure, and funnel structure. 

\paragraph{DoE2Vec features learned using a variational autoencoder} DoE2Vec features~\cite{doe2vec} are low-level features learned in an unsupervised manner with a variational autoencoder (VAE)~\cite{kingma2019introduction}. To learn them, first, a dataset of candidate solutions, initially within the domain of [0,1]$^d$ ($d$ is the problem dimension) is generated using a sampling technique. Next, the candidate solutions are evaluated using a set of problem instances that are randomly generated using the random function generator from~\cite{long2022learning}, which is a modification of the generator proposed in~\cite{tian2020recommender}. This evaluation allows the collection of training data with different complexity that covers a wide range of problem instances. Further, all objective solution values are first re-scaled within the range of [0,1] and are then used as input features to train the VAE. These features not only demonstrate promise in identifying similar surrogate problem instances that are inexpensive to evaluate but also have the potential to significantly enhance performance when used alongside traditional ELA features in problem instance classification tasks.

\paragraph{TransOpt features learned using a transformer} TransOpt features~\cite{cenikj2023transopt} are low-level features learned in a supervised manner by using a transformer architecture~\cite{vaswani2017attention}. The original transformer architecture's core lies in its encoder-decoder structure, comprising multiple identical blocks for both the encoder and decoder. In the case of TransOpt features, only the encoder has been utilized, consisting of a multi-head self-attention module and a position-wise feed-forward network. The input data for training the transformer are generated candidate solutions for each problem instance, obtained using Latin Hypercube sampling. These samples are provided to the transformer encoder, which produces a representation for each sample. A representation of the problem is then obtained by aggregating the sample representations.

This architecture has been trained in a supervised manner on two different tasks. In~\cite{cenikj2023transopt}, the model is trained on the BBOB benchmark, where the task is to identify to which of the 24 BBOB problem classes an instance belongs. In this case,
 24 problem classes with 999 instances (i.e., shifted, scaled) per each problem class and per dimension $d \in \{3, 5, 10, 20\}$ are used. The tested sample sizes are 50$d$ and 100$d$. The model consists of the previously mentioned encoder and a classification head trained to predict the 24 problem classes. 
 It has been shown that TransOpt features can achieve accuracy rates ranging from 70\% to 80\% when tasked with identifying problem classes across various problem dimensions. It is important to note that even though the model is trained on the problem classification task, it can produce feature representations for unseen problem instances which can be used in different downstream tasks. A demonstrative example of this is shown in~\cite{gina_generalization_study}, where the TransOpt model is used to generate problem representations on top of which a random forest model~\cite{biau2016random} is trained for the algorithm selection task.

\paragraph{Deep-ELA features learned using a transformer} Deep-ELA~\cite{seiler2024deep} is a methodology involving the unsupervised training of transformer models to produce representations of optimization problems which are invariant to problem transformations. This is accomplished by training the model to produce the same representation for a problem instance and its augmented variant. The augmentation is accomplished through transformations that do not alter the underlying optimization problem, such as rotations and inversions of decision variables and randomization of the sequence of decision and objective variables. 

Four transformers have been pre-trained on large sets of randomly generated optimization problems to grasp intricate representations of continuous single- and multi-objective landscapes. The Deep-ELA features have been evaluated in the problem classification task classifying the BBOB single-objective problem based on their high-level characteristics defined in the benchmark suite. In addition, they have been evaluated in automated algorithm selection tasks on the BBOB benchmark suite and four different multi-objective benchmark suites. A study evaluating the complementarity of Deep-ELA and classical ELA features was recently presented in~\cite{deep_classical_LION}, with a following study involving TransOpt features~\cite{seiler2024learned}. 

\paragraph{Random Filter Mappings}
As opposed to using trained deep learning models to extract problem landscape features,~\cite{random_filters_gasper} proposes the extraction of features using randomly initialized filters. The procedure starts with the initialization of a set of filters with random sizes, weights and radii. Each individual filter is then applied at
randomly selected anchor points (samples of the objective function). In particular, the filter application involves the multiplication of the filter weights with the pairwise distances of a subset of the problem samples which are in the vicinity of the anchor point (have a distance which is less than the dimension-adjusted radius of the filter). The thereby obtained filter responses are aggregated to obtain the problem landscape features. 
These features have been evaluated for the algorithm selection task, as well as the task of problem classification of the BBOB problem instances into the 24 problem classes, and recognizing their high-level properties.

\subsection{Advantages vs. Disadvantages of Problem Landscape Features}
After summarizing the latest trends in problem landscape features, Table~\ref{tab:pvc_problem} presents their advantages and disadvantages.
\begin{table}[!htp]\centering
\caption{Advantages vs. disadvantages of the different problem landscape feature sets.}\label{tab:pvc_problem}
\scriptsize
\begin{tabular}{lp{5cm}p{5cm}}\toprule
\textbf{Problem landscape features} &\textbf{Advantages} &\textbf{Disadvantages} \\\midrule
\multirow{4}{*}{ELA} 
    &- Easy use through R, Python, and GUI 
    &- Sensitive to sampling strategy and sample size \\
    &- Most commonly-used problem landscape features 
    &- Noninvariant to problem shifting, scaling, and rotation \\
    &- Explainable to some degree 
    &- Poor generalization on new unseen instances when plugged into an ML model \\
    & &- Computationally costly \\
    \hline
TLA 
    &Invariant to shifting, scaling, and rotation 
    &- Black-box features \\
    & &- Can produce a very high-dimensional feature vector, may require dimensionality reduction before usage \\
    \hline
\multirow{2}{*}{Learned by CNNs} 
    &- More applicable for two and three dimensional problems 
    &- Black-box features \\
    &- Does not require any feature engineering 
    &- Loss of high-dimensional information by representing the data as a two dimensional fitness map \\
    & &- Depends on the problem portfolio used for learning \\
    \hline
\multirow{2}{*}{\shortstack[l]{Learned by point\\ cloud transformer}} 
    &- Applicable to all problem dimensions, though may require model retraining for higher dimensions 
    &- Black-box features \\
    & &- Depends on the problem portfolios used for learning \\
    \hline
Doe2Vec 
    &- Does not require any feature engineering 
    &- Black-box features \\
    \hline
\multirow{3}{*}{TransOpt} 
    &Does not require any feature engineering 
    &- Black-box features \\
       \hline
Deep-ELA 
    &- Does not require any feature engineering 
    &- Black-box features \\
    \hline
\multirow{2}{*}{Random filter mappings} 
    &- Does not require any feature engineering 
    &- Black-box features \\
    & &- May produce irrelevant features \\
\bottomrule
\end{tabular}
\end{table}

\subsection{Summary of Problem Landscape Features}

Table~\ref{tab:sum_problem_landscape} demonstrates a summary of the type of training tasks and inputs required for each problem landscape feature group. Additionally, the last column contains information about how the features are calculated in different studies. In particular, we report the size of the sample used for feature calculation, the number of repetitions performed when calculating the features (since some of them are stochastic and some studies perform multiple runs and aggregate the feature values), and the dimension of the problems on which the features are tested. Please note that if the study did not explicitly report that multiple repetitions were performed, we assume that the features were calculated once.
Furthermore, please note that for the ELA features, we do not include an exhaustive list of studies where the features were used, due to an overwhelmingly high number of such studies. Instead, we chose 10 studies from different authors as examples.
From the table, we can see that there is a lack of consistency in feature calculation.
First, there is substantial variation in sample sizes between studies. Similarly, the number of repetitions varies widely: some studies rely on a single run, while others average results across 10, 15, or even 100 repetitions to mitigate stochastic variability.
Another clear observation is that most empirical work has been conducted on low-dimensional problems, typically up to $d=10$ or $d=20$, with a few exceptions exploring higher dimensions. This suggests that while problem landscape features have been extensively explored in principle, their evaluation has been constrained to relatively small search spaces.
This highlights the need for systematic evaluations of feature robustness in higher-dimensional settings, where sampling becomes more expensive but also more informative.

Future work should test how features scale across different dimensionalities, analyze their stability across varying sample sizes, and assess sensitivity to function transformations, stochasticity and repetition strategies. Moreover, standardized protocols are needed on \textit{i)} how many samples to draw per dimension, \textit{ii)} how many repetitions are required for statistically reliable estimates, and \textit{iii)} which dimensions should be included in benchmarking studies. While some studies have already addressed the sensitivity of ELA features to function transformations~\cite{urban_ela_not_invariant}, sampling strategies and sample sizes~\cite{lang2021exploratory, urban_ela_not_invariant_sampling}, these questions remain relatively underexplored for more recent feature groups.

\begin{table}[!htp]
\centering
\caption{Summary of inputs, training task types and sample sizes used in the literature for problem landscape features}\label{tab:sum_problem_landscape}
\scriptsize
\begin{tabular}{p{1.5cm}p{1.5cm}p{3cm}p{7cm}}
\toprule
\textbf{Feature name} & \textbf{Training task type} & \textbf{Inputs} & \textbf{Sample} \\
\midrule
ELA & manually designed & Samples of optimization problems & 
- 250$d$ samples; 100 repetitions; $d \in \{5\}$ ~\cite{eftimov2020linear}

- 500$d$ samples; 10 repetitions; $d \in \{2\}$ ~\cite{ela_cell_mapping}

- 250$d$ samples; 10 repetitions; $d \in \{10\}$~\cite{skvorc2022transfer}

- 6$d$, 60$d$, 625$d$ samples; $d \in \{5\}$~\cite{renau2020exploratory}

- 1000$d$ samples; 15 repetitions; $d \in \{2,3,5,10,20\}$ ~\cite{munoz2012meta}

- 1000$d$ samples; 1 repetition; $d \in \{2,5\}$~\cite{MABBOBteloversion}

- 50$d$, 100$d$ samples; 1 repetition; $d \in \{3,10\}$~\cite{gina_generalization_study}

- 250$d$ samples; 1 repetition; $d \in \{5\}$~\cite{gasper_pitfalls}

- 50$d$, 100$d$ samples; 10 repetitions; $d \in \{2,3,5,10\}$~\cite{pascal_low_budget_ela}

- 250$d$ samples; 1 repetition; $d \in \{5, 10\}$~\cite{kudela2023computational}

\\ \midrule
TLA & manually designed & Samples of optimization problems & 
- 10, 40, 100, 200, 400, 600 samples; 1 repetition; $d \in \{2,3,5,10\}$~\cite{tinytla}

- 50$d$ samples, 1 repetition, $d \in \{10\}$ \cite{gina_hit_the_wall}\\ \midrule
Fitness Map + CNNs & supervised & Samples of optimization problems, High-level problem properties & - 100, 500 samples; 10 repetitions; $d \in \{2,3,5,10\}$~\cite{seiler2022collection}\\ \midrule
Point Cloud Transformer & supervised & Samples of optimization problems, High-level problem properties & - 100, 500 samples; 10 repetitions; $d \in \{2,3,5,10\}$~\cite{seiler2022collection}\\ \midrule
DoE2Vec & self-supervised & Samples of optimization problems & 
- 256 samples; 1 repetition; $d \in \{5\}$~\cite{doe2vec}

- 50$d$ samples; 1 repetition; $d \in \{10\}$problems~\cite{gina_hit_the_wall} \\ \midrule
TransOpt & supervised & Samples of optimization problems, Problem class & 
- 50$d$, 100$d$ samples; 1 repetition; $d \in \{3,20\}$~\cite{cenikj2023transopt}

- 50$d$, 100$d$ samples; 1 repetition; $d \in \{3,10\}$~\cite{gina_generalization_study}

\\ \midrule
Deep-ELA & self-supervised & Samples of optimization problems & 
- 25$d$, 50$d$ samples; 1 repetition; $d \in \{2,3,5,10\}$~\cite{seiler2024deep}

- 50$d$ samples; 1 repetition; $d \in \{10\}$ \cite{gina_hit_the_wall} 

- 25$d$, 50$d$ samples; 1 repetition; $d \in \{2,3,5,10\}$~\cite{deep_classical_LION}

- 50$d$ samples; 1 repetition; $d \in \{3,10\}$~\cite{seiler2024learned}
\\ \midrule

Random Filter Mappings & manually designed & Samples of optimization problems & - 200$d$ samples; 1 repetition; $d \in \{2,3,5,10\}$~\cite{random_filters_gasper}\\ \midrule
\end{tabular}
\end{table}

\section{Algorithm Features}
\label{sec:algorithm_representation}
Algorithm representations are focused on calculating features that characterize the algorithm instance. While the development of algorithm features has garnered somewhat lower research interest compared to the development of problem landscape features, in this section we present the features derived directly from the source code.

\paragraph{Algorithm Features Based on Source Code}
The extraction of features from the algorithm source code and from the abstract syntax tree has been proposed in~\cite{algorithm_code_features}. The features extracted from the source code capture the cyclomatic complexity, maximum indent complexity, number of lines of code, and size in bytes for the entire source code as well as aggregations of these properties across different regions of the code. On the other hand, the features extracted from the abstract syntax tree involve the conversion of the abstract syntax tree obtained during the compilation of a given algorithm into a graph representation and extracting various graph properties such as node count, edge count, transitivity, node degree, clustering coefficient, depth, etc. 

In addition to source code-derived metrics, potential algorithm representations could also capture functional and structural characteristics that directly reflect the design choices of metaheuristics. For example, they could include the types of variation operators employed (e.g., mutation, crossover, or recombination), the selection and replacement mechanisms (e.g., tournament, rank-based, elitist strategies), and the presence of restart or perturbation policies. 
Another important dimension is parameter adaptivity. Algorithms may embed self-adaptive mechanisms (e.g., step-size control in evolution strategies, cooling schedules in simulated annealing) or hyper-heuristic controllers that dynamically adjust parameters during the search. Representing whether and how such adaptivity is employed could offer valuable insights into algorithm behavior. Please note that these are potential research directions and not existing works.

\subsection{Advantages vs. Disadvantages of Algorithm Features}
 Table~\ref{tab:pvc_algorithm} presents their advantages (pros) and disadvantages (cons) of the algorithm features based on the source code.

\begin{table}[!htp]\centering
\caption{Advantages vs. disadvantages of algorithm features.}\label{tab:pvc_algorithm}
\scriptsize
\begin{tabular}{lp{5cm}p{5cm}}\toprule
\textbf{Algorithms features } &\textbf{Advantages} &\textbf{Disadvantages} \\\midrule
\multirow{2}{*}{Based on source code} 
    &- May be used to compare different programming implementations of the algorithms and further investigate which one has better performance 
    &- These features are ineffective for automated algorithm configuration or parameter tuning, as parameter differences are typically evident only during execution, not in the code. \\
    & &- Features extracted from the source code depend highly on the programming language and the specific implementation, leading to potential discrepancies even for the same algorithm. \\
\bottomrule
\end{tabular}
\end{table}

\section{High-Level Problem-Algorithm Interaction Features}
\label{sec:high-level-problem-algorithm_representation}
\begin{figure}
    \centering
    \includegraphics[width=1\linewidth]{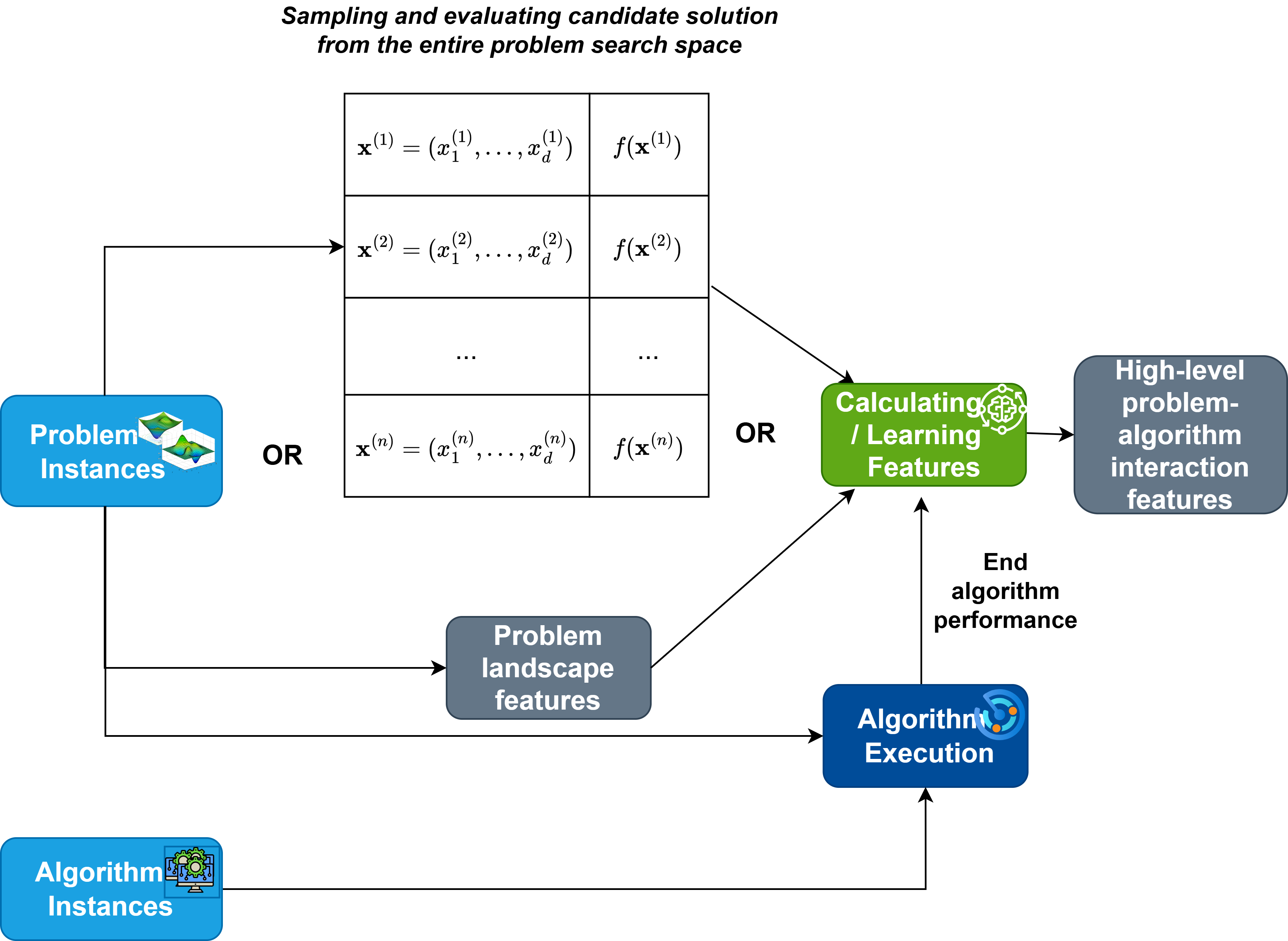}
    \caption{The general pipeline for calculating high-level problem-algorithm interaction features.}
    \label{fig:pip_problem-algorithm}
\end{figure}

These features are based on general characteristics of the optimization problems where the algorithms are applied, instead of describing how the algorithm’s search process unfolds over time. Essentially, they establish a link between the problem’s high-level properties and the algorithm’s final performance, but they do not capture details of the algorithm’s behavior during its run. Figure~\ref{fig:pip_problem-algorithm} presents the general pipeline of calculating high-level problem-algorithm landscape features.

\paragraph{Features learned by using convolutional neural networks} CNN-based features were investigated in~\cite{prager2021towards,seiler2022collection}, and their details were previously discussed in Section~\ref{sec:problem_landscape_features}. The key difference here lies in the supervised learning task. For problem landscape features, the supervised task involved classifying problem instances based on high-level properties without considering any algorithm interaction. In contrast, in this case, the model was trained for algorithm selection, aiming to choose the most suitable modular CMA-ES~\cite{HansenO01} configuration from a portfolio of 32 for each problem instance. The latent representations obtained from the CNN are subsequently used as low-level features.

\paragraph{TransOptAS features learned using a transformer} Similar to TransOpt features presented in Section~\cite{cenikj2023transopt}, where the transformer has been trained in a problem classification task, in~\cite{transoptas}, the transformer architecture is directly trained on the algorithm selection task. In this case, the model consists of an encoder and a regression head that predicts a numerical performance indicator for 12 configurations of the PSO algorithm.

\paragraph{Features Based on Performance}
\label{sec:p2v}
Performance2Vec~\cite{performance2vec} is a methodology for constructing features that describe problem-algorithm interaction based on the performance achieved on a set of benchmark problems. Here, the vector representation consists of the algorithm instance performance obtained on each benchmark problem separately, so they can be assumed to be different features. Algorithms' performance can be defined in terms of different metrics, using simple statistics like mean or median, or more complex ranking schemes like Deep Statistical Comparison~\cite{dsc}.

\paragraph{Features Based on Problem Landscape Features used by Performance Prediction Models}
\label{sec:shap}
This type of representation is generated by observing the importance assigned to ELA features by explainability methods applied to ML models trained for performance prediction. It involves the use of the SHAP~\cite{shap} method, which can assign feature importance on a \textit{global} level (i.e., on a set of problem instances) and on a \textit{local} level (i.e., on a particular problem instance). A single-target regression (STR) model is trained to predict the performance of the algorithm instance using the training problem instances described by their ELA features. The model is then used for making predictions on the problem instances from the test dataset. Next, using the SHAP method, the Shapley values that are the contributions of each landscape feature to the performance prediction are calculated. If the input to the model are $p$ landscape features, averaging the Shapley values across all problems for each landscape feature separately gives a $p$-dimensional meta-representation of the algorithm behavior. This type of representation encodes the interactions between the landscape features and the algorithm performance. Such representations have been previously used to find algorithm instances with similar behavior on a set of benchmark problem instances~\cite{anan_ela_feature_importance_2} and to understand which module from a modular CMA-ES~\cite{HansenO01} is active based on the performance data~\cite{anak_ela_feature_importance}.

\paragraph{Features Using Graph Embeddings}
Features can also be constructed by leveraging their interactions with different entities from the optimization domain. In~\cite{anak_kg_performance_prediction}, the performance prediction task was addressed as a link prediction task, which aims to identify if an algorithm can solve a given problem instance with some specified error. In this case, the methodology involved the construction of a knowledge graph (KG), where some of the nodes describe aspects of the problem instances (their corresponding problem class, high-level features, and ELA features), while other nodes are related to algorithm descriptions (their parameters). The algorithm and problem instance nodes are linked if the algorithm can solve the problem instance with some specified error threshold. Such a representation of the algorithm and problem instance properties allows the application of standard knowledge graph embeddings for deriving representations of the problem instances and algorithm instances, which can then be used for performance prediction. We need to highlight here that this approach of learning features can produce either algorithm features if the node is representing an algorithm instance or problem instance features if the node is representing a problem instance. In the case of problem instance features, they differ from the low-level landscape features described in Section~\ref{sec:problem_landscape_features} since they fuse information not only for the landscape but also consider the interaction of an algorithm instance on each problem instance through its end performance.

\paragraph{Features Using Graph Neural Network}  Parameters describing algorithm configurations, which significantly influence performance, were often overlooked by the techniques of learning algorithm features. To address this, the complex structure formed by the relationships between algorithm operators/modules, parameters, problem characteristics, and performance outcomes was represented as a graph~\cite{kostovska2025geometric}. Heterogeneous graph data structures and graph neural networks were explored to enable the prediction of optimization algorithm performance by capturing the complex dependencies between problems, algorithm configurations, and performance outcomes. Two modular frameworks, modCMA-ES and modDE, were considered. Improvements of up to 36.6\% in mean square error in algorithm performance prediction over traditional tabular-based methods were achieved, and the potential of geometric learning in black-box optimization was demonstrated. This approach generates an algorithm and problem landscape features depending on the node of interest.

\paragraph{Features based on Configuration Settings for Modular Algorithm Frameworks} 
Another approach for extracting features from algorithm configurations involves using the importance or contribution of operators and modules to the performance of the algorithm configuration. Two approaches are presented here:
\begin{itemize}
    \item Features based on functional ANOVA (fANOVA) - Problem class-specific datasets are initially generated, in which modules are used as features of the modular algorithm variants, with numerous variants incorporating different module values being executed~\cite{nikolikj2024quantifying}. The corresponding performance on the problem class is treated as the target within each dataset, thereby capturing the relationship between the algorithmic modules and performance in that specific problem class. These datasets are then provided as input to f-ANOVA~\citep{hutter2014efficient}, through which the variance in performance is decomposed and systematically attributed to individual modules and their combinations. When variants are defined by $m$ modules, the analysis quantifies the effect (i.e., importance) on performance for $m$ individual modules, $\binom{m}{2}$ pairwise module interactions, and $\binom{m}{3}$ triple module interactions. Estimation of higher-order interactions is considered computationally expensive. The complete set of effects (singles, pairs, and triplets) is subsequently used as a feature vector, and by comparing these vectors across different problems, problems can be identified in which the modular frameworks exhibit similar module interactions that describe their performance.
    \item Features based on Shapley analysis - Similar to features generated by fANOVA, Shapley analysis can also be used to estimate the contribution of each module to performance~\cite{vanstein2024explainable}. However, a limitation of this approach is that only the contributions of individual modules are calculated, as estimating higher-order contributions is computationally expensive.
    
\end{itemize}

\subsection{Advantages vs. Disadvantages of High-Level Problem-Algorithm  Interaction Features}
After summarizing the latest trends in high-level problem-algorithm  interaction features, Table~\ref{tab:pvc_algorithm-ptoblem-high} presents their advantages (pros) and disadvantages (cons).

\begin{table}[!htp]\centering
\caption{Advantages vs. disadvantages of high-level problem-algorithm interaction features.}\label{tab:pvc_algorithm-ptoblem-high}
\scriptsize
\begin{tabular}{lp{5cm}p{5cm}}\toprule
\textbf{High-Level p-a features } &\textbf{Advantages} &\textbf{Disadvantages} \\\midrule
\multirow{2}{*}{Learned by CNNs} 
    &- More applicable for two-dimensional and three-dimensional problems 
    &- Black-box features \\
    &- Does not require any feature engineering 
    &- Loss of high-dimensional information by representing the data as a two-dimensional fitness map \\
    & &- Depends on the algorithm portfolio used for learning \\
    \hline
    \multirow{3}{*}{TransOptAs} 
    &- Does not require any feature engineering 
    &- Black-box features \\
    & &- Depends on the algorithm portfolio used for learning \\
    & &- Requires algorithm execution for obtaining algorithm performance labels used for supervised training \\
    \hline
Based on performance 
    &- Facilitates algorithm comparison through performance vectors 
    &- Biased to the selected portfolio of benchmark problems \\
    \hline
\multirow{3}{*}{Based on Shapley values} 
    &- Encodes interactions between problem landscape features and algorithm performance 
    &- Depends on the selected problem landscape features portfolio \\
    &- Used to find similar algorithm behaviors with the assumption that the predictive models behave similarly 
    &- Depends on the selected benchmark problem instances \\
       & &- The dependencies on the surrogate (STR) model \\
  
    \hline
 via Knowledge Graph 
    &- Encodes interactions between problem landscape features and algorithm performance by also involving the graph neighborhood 
    &- Depends on the data stored in the KG and the KG embedding method used for learning\\
    \hline
via GNNs 
    &- Encodes interactions between problem landscape features, algorithm configuration, and algorithm performance by involving the graph neighborhood 
    &- Depends on the data stored in the graph and the GNN method used for learning\\
    \hline
    Based on fANOVA 
    &- For each problem, they encode the contributions of modules and their interactions to the final performance of an algorithm.
    &- Depends on the available data from the pool of different configurations. \\
    \hline
    Based on SHAP 
    &- For each problem, they encode the contributions of each separate module to the final performance of an algorithm.
    &- Depends on the available data from the pool of different configurations. \\  
\bottomrule
\end{tabular}
\end{table}

\subsection{Summary of High-Level Problem Algorithm Interaction Features}

Table~\ref{tab:sum_high_level} provides a structured overview of the high-level problem-algorithm interaction features studied in the literature, focusing on their training task type, the inputs required, and the sample sizes reported. Unlike problem landscape features, which primarily rely on sampled candidate solutions evaluated on the objective function, these interaction features explicitly incorporate information about algorithm behavior and performance.

The table illustrates that a variety of training paradigms are used. 
With respect to inputs, these features require richer information than problem landscape features. For example, Performance2Vec uses only algorithm performance vectors, whereas KG and GNN embeddings combine algorithm performance, ELA features, and configuration parameters. This integration makes them more expressive but also more expensive to compute, as they often depend on preliminary feature calculations or repeated algorithm runs.

\begin{table}[!htp]
\centering
\caption{Summary of inputs, training task type and sample sizes used in the literature for high-level problem-algorithm interaction features}\label{tab:sum_high_level}
\scriptsize
\begin{tabular}{p{2cm}p{1.5cm}p{3cm}p{7cm}}
\toprule
\textbf{Feature name} & \textbf{Training task type} & \textbf{Inputs} & \textbf{Sample} \\
\midrule

Fitness Map + CNNs & supervised & Samples of optimization problems, algorithm performance & - 50$d$ samples; 1 repetition; $d \in \{2\}$ ~\cite{prager2021towards} \\\midrule
TransOptAS & supervised & Samples of optimization problems, algorithm performance & - 50$d$ samples; 1 repetition ; $d \in \{3,10\}$~\cite{transoptas}
- 50$d$ samples; 1 repetition ; $d \in \{10\}$~\cite{gina_hit_the_wall}
\\\midrule
Performance2Vec & manually designed & Algorithm performance & - No samples used; $d \in \{10\}$~\cite{performance2vec}  \\\midrule
Explainable Prediction Models & supervised & Algorithm performance, calculated ELA features & - 800$d$ samples; 30 repetitions; $d \in {10}$~\cite{anan_ela_feature_importance_2} \\ \midrule
Internal Algorithm Parameters & unsupervised & Time series of algorithm parameters during runs & - No samples used; $d \in \{5\}$~\cite{algorithm_features_time_series_cmaes}\\ \midrule
KG embeddings & self-supervised & Algorithm performance, calculated ELA features, algorithm linked with configuration parameters & 
- 100$d$ samples; 100 repetitions; $d \in \{5,30\}$ (for ELA feature calculation) ~\cite{anak_kg_performance_prediction}
\\ \midrule
GNN embeddings & supervised & Algorithm performance, calculated ELA features, algorithm linked with configuration parameters & - 100$d$ samples; 100 repetitions; $d \in \{5,30\}$ (for ELA feature calculation)~\cite{kostovska2025geometric} \\ \midrule
fANOVA & manually designed & Algorithm performance & - No samples used; $d \in \{5,30\}$~\cite{nikolikj2024quantifying}\\ \midrule
\end{tabular}
\end{table}

\section{Trajectory-based features that capture problem-algorithm interaction}
\label{sec:trajectory_based_representation}
 
\begin{figure}
    \centering
    \includegraphics[width=1\linewidth]{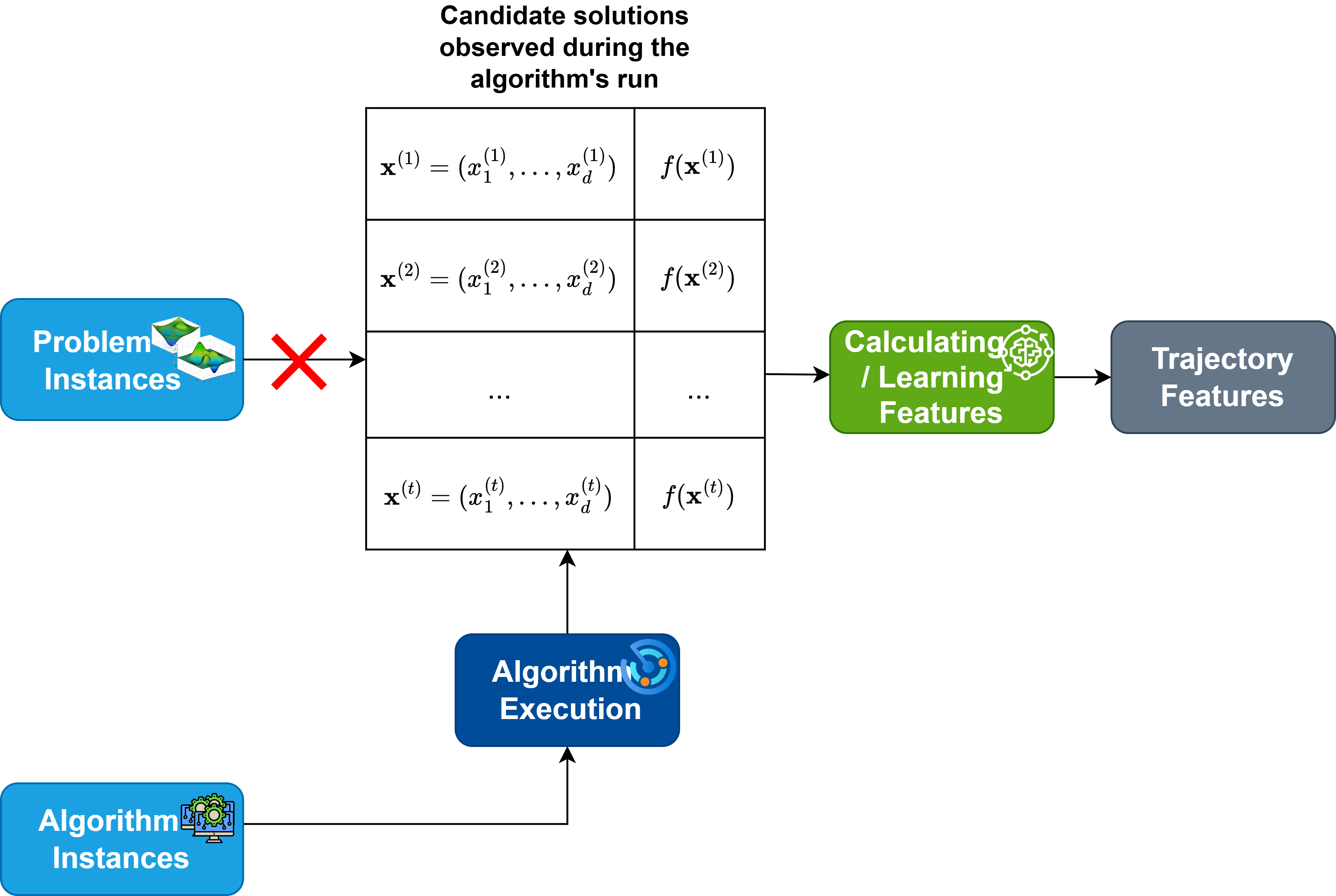}
    \caption{The general pipeline of calculating problem-algorithm trajectory features.}
    \label{fig:pip_trajectory}
\end{figure}

The previously mentioned approaches for problem landscape features compute low-level features by generating an artificial set of candidate solutions using a sampling strategy that spread well across the whole decision space. 
They are hence not connected to the actual set of candidate solutions explored by an optimization algorithm instance \textit{during} its optimization process. To address this limitation, several studies have undertaken automated algorithm selection on a per-instance and per-run basis by calculating \textit{trajectory-based features} that represent the algorithm behavior during its run. Below, we are going to provide more details about different trajectory-based approaches for learning features that capture problem-algorithm interaction. Figure~\ref{fig:pip_trajectory} presents the general pipeline of calculating problem-algorithm trajectory features.

\paragraph{Trajectory-based features Based on Internal Algorithm Parameters}

Some optimization algorithms contain internal parameters that are adapted throughout the optimization process, such as the step size, the evolution path, the conjugate evolution path, and the square root of the diagonal of the diagonal matrix holding the eigenvalues of the covariance matrix. Building on features suggested in~\cite{HolenaCMAfeaturesGECCO19}, the work~\cite{algorithm_features_time_series_cmaes} studied the use of time-series features (using the \textit{tsfresh}~\cite{tsfresh} library) on top of these algorithm-internal parameters. These time series features were used to identify the configuration of modular CMA-ES variants.

\paragraph{Trajectory-based ELA features} 
Trajectory-based ELA features are computed from a set of candidate solutions that are derived from the entire algorithm trajectory, rather than from a standard sampling technique (such as Latin Hypercube Sampling~\cite{lhs}, random sampling, etc.) that is applied to the objective function. This means that the samples are extracted from the populations (i.e., sets of candidate solutions and their corresponding function values) that are generated/observed by the optimization algorithm during its execution. The trajectory-based ELA features have been first proposed in~\cite{adaptive_landcape_analysis}, under the name of \textit{Adaptive Landscape Analysis}. In that study, the ELA features are computed using samples from the distribution that the algorithm CMA-ES uses to sample its solution candidates. The approach has been evaluated for fixed-budget performance prediction~\cite{trajectory_ela_performance_regression} of the CMA-ES~\cite{HansenO01}, as well as \textit{per-run} algorithm selection with warm-starting~\cite{per_run_algorithm_selection_warmstarting}, where the features extracted from the trajectory of an initial optimization algorithm instance are used to determine whether to switch to a different algorithm instance. We need to highlight here that the trajectory-based ELA features use a subset of all candidate solutions explored by the algorithm instance during its run, without considering the iteration in which these solutions were generated. These features do therefore not capture the longitudinality of the solutions that are observed within the iterations of the algorithm execution. Another study that uses the trajectory-based ELA has been conducted to analyze the benefits of predicting the switching between different algorithm instances during the optimization trajectory run~\cite{vermetten2023switch}. It has been also shown that they can be used for dynamic selection of the acquisition function of Bayesian Optimization~\cite{bayesian_optimization} algorithms, improving over default static choices~\cite{benjamins2022towards}.

\paragraph{DynamoRep features} DynamoRep features~\cite{10.1145/3583131.3590401, dynamorep} are capturing the problem-algorithm interaction using simple descriptive statistics extracted from the populations explored by the algorithm in each iteration of its execution on a problem instance. In particular, the minimum, maximum, mean, and standard deviation of each of the candidate solutions and the objective function values of each population are extracted. A representation of the entire trajectory is then generated by concatenating the values of these statistics extracted for each population.
If the algorithm instance is run for $b$ iterations until the stopping criteria are met on a problem instance of dimension $d$ (i.e., number of decision variables), the entire algorithm trajectory representation would then have a size of $4b(d + 1)$. The +1 is coming from the objective function value. DynamoRep features have been used for algorithm selection, problem classification and algorithm classification~\cite{dynamorep}.

\paragraph{Opt2Vec features} The trajectory-based ELA and DynamoRep features are designed to analyze and represent the entire trajectory of an optimization run. However, an important aspect is the information contained within a small, specific segment of the optimization trajectory, such as a particular timestamp, iteration, or population. This is especially crucial when it comes to optimizing dynamic algorithms efficiently. For this purpose, 
 Opt2Vec~\cite{korovsec4316939opt2vec} features propose the usage of autoencoders to encode the information found in the interaction between an optimization algorithm and optimization problem into an embedded subspace. These representations take the populations considered by an optimization algorithm instance in each iteration of its execution, scale the candidate solutions and objective function values, and identify a subset of informative populations. The subset of informative populations is obtained by calculating the Frobenius norm on the differences of the matrices representing consecutive populations of the algorithm's execution, and discarding one of the populations if the Forbenius norm is smaller than a predefined threshold. The proposed approach is designed to capture features for parts of the optimization problem's search space that the algorithm explores at a specific timestamp or iteration during the optimization process. The Opt2Vec features are evaluated for the task of classifying problems from the CEC benchmark suite. In particular, 30 CEC problems in three different dimensions (10,30,50) are used. Different dimensionalities of the same problem are treated as different problems, resulting in 90 problem classes used in the classification task. The Opt2Vec features achieve classification accuracies over 80\% for the recognition of the problem being solved and its dimension.
 
\paragraph{Iterative-based ELA features} An indirect outcome of the Opt2vec and DynamoRep studies are the iterative-based ELA features, which involve calculating the ELA features on each population separately, and concatenating the population representations to generate a representation for the entire trajectory. In this way, the representations would capture the dynamics and behavior of the algorithm on the problem landscape, however, the dimension of the generated representation (i.e., number of features) grows linearly with the number of iterations for which the algorithm is run. 
The iterative ELA features have been evaluated for algorithm selection, problem classification and algorithm classification~\cite{dynamorep, opt2vec}.

\paragraph{Local Optima Networks} Local Optima Networks (LONs)~\cite{ochoa2014local} serve as a simplified model for representing discrete (combinatorial) fitness landscapes, where local optima are nodes and search transitions are edges defined by an exploration search operator. They represent the number of local optima in the landscape, as well as their distribution and connectivity patterns.
Monotonic LONs (MLONs) are a variant of LONs where transitions between local
optima are considered only if fitness is non-deteriorating. To account for neutrality in the level of local optima transitions, i.e., connected components in the MLONs that have the same fitness value, 
compressed MLONs (CMLONs) have been proposed~\cite{adair2019local}. A compressed local optimum can be described as a grouping of interconnected nodes within the MLONs, all sharing the same fitness value. 
While originally proposed for combinatorial problems, LONs have been adapted for the continuous domain in~\cite{adair2019local}. In this case, a Basin-Hoping algorithm~\cite{basin_hopping} is used to identify the local optima, and the MLONs and CMLONs are adapted for the continuous problems.

CMLONs have been utilized to visualize the 24 BBOB problem classes across different problem dimensions~\cite{mitchell2023local}. Network metrics are also calculated for each CMLON, and dimensionality reduction techniques are employed to classify and compare these problems. The findings reveal that CMLONs exhibit varied representations that are linked to the inherent properties of the problem instances and their dimensionality. Network metrics prove particularly crucial for multimodal problems in higher dimensions, where CMLONs become too intricate for meaningful visual interpretation.

Similar to LONs are the Search Trajectory Networks (STNs)~\cite{ochoa2020search}, which are novel tools to study and visualize how population-based algorithm instances behave in continuous spaces. Inspired by LONs, which map the nodes to local optima, in STNs the nodes focus on different states from the optimization trajectory not limited to local optima. The edges signify the progression between these states. This expansion enhances network-based models' utility for understanding heuristic search algorithms. However, they have not yet been evaluated as possible sources for learning trajectory-based features.

\paragraph{Probing trajectories} 
In a recent study, researchers introduced an algorithm selector method that utilizes short probing trajectories generated by running an initial algorithm on a problem instance for a brief period~\cite{renau2024utility}. These trajectories are then used to determine either the current fitness or the best-so-far fitness, which are tracked across predefined sequential iterations of the initial algorithm. Following this, time-series features are extracted using the \textit{tsfresh} library. These time-series features are then used as input for a Random Forest classifier~\cite{biau2016random} to train the algorithm selector. Additionally, the raw probing trajectories are fed into a Rotation Forest classifier for the same purpose. This approach shows promising results comparable to those achieved using trajectory-based ELA features extracted from observed candidate solutions and a Random Forest model. 

\paragraph{ClustOpt}  A representation learning and visualization methodology is proposed~\cite{cenikj2025clustopt}, in which solution candidates explored by the algorithm are clustered, and the evolution of cluster memberships across iterations is tracked. In this way, a dynamic and interpretable view of the search process is provided. Additionally, two metrics—algorithm stability and algorithm similarity—are introduced to quantify the consistency of search trajectories across runs of an individual algorithm and the similarity between different algorithms, respectively. This methodology is applied to a set of ten numerical metaheuristic algorithms, through which insights into their stability and comparative behaviors are revealed, thereby enabling a deeper understanding of their search dynamics.

\subsection{Advantages vs. Disadvantages of Trajectory Features.}
After summarizing the latest trends in problem-algorithm trajectory features, Table~\ref{tab:pvc_trajectory} presents their advantages and disadvantages.
\begin{table}[!htp]\centering
\caption{Advantages vs. disadvantages of trajectory features}\label{tab:pvc_trajectory}
\scriptsize
\begin{tabular}{lp{5cm}p{5cm}}\toprule
\textbf{Trajectory features } &\textbf{Advantages} &\textbf{Disadvantages} \\\midrule
Based on internal parameters 
    &- Capture the behavior of the algorithm 
    &- Lack of comprehensive comparison of different time series features \\
    &
    &- Do not enable a comparison of different algorithms (with different configurations), limited to comparing configurations of the same algorithm \\
    \hline
Trajectory-based ELA 
    &- Info about the interaction across problem and algorithms (personalization) 
    &- Does not capture the longitudinal aspect of solutions within algorithm iterations \\
    \hline
Iterative-based ELA 
    &- Info about a single timestamp of the optimization process, can easily be combined with ML models that will capture the longitudinality of the search process
    &- ELA features are sensitive on small sample sizes \\
    \hline
\multirow{2}{*}{DynamoRep} 
    &- cheap to compute  
    &- Limited expressiveness \\
    &- Despite lower computational cost, DynamoRep features yield similar performance as ELA features that are calculated at each iteration of the algorithm's execution 
    &- Representation size grows with number of iterations and problem dimension, may require dimensionality reduction as preprocessing step \\
    \hline
\multirow{3}{*}{Opt2vec} 
    &- Capture features specific to parts of the search space explored at a particular iteration 
    &- Depends on the data used to train the autoencoder \\
    &- Takes into consideration the optimization problem dimension since one model is used for all problem dimensions.
    &- Black-box features \\
    \hline
LON and variants 
    &- Useful for visualization purposes 
    &- Very costly to compute \\
    \hline
Probing trajectories 
    &- Potential to be utilized for per-run algorithm selection 
    &- Requires suitably chosen probing algorithms \\
    & & - Difficult to transfer results from one probing algorithm to another\\
    \hline
   ClustOpt
    &- Useful for algorithm behavioral analysis
    &- Depends on the clustering algorithm and its parameters \\
\bottomrule
\end{tabular}
\end{table}

\subsection{Summary of Trajectory Features}

Table~\ref{tab:sum_trajectory} summarizes the trajectory-based feature families explored in the literature, highlighting their training task type, the inputs required, and the problem dimensions on which they have been tested. Unlike problem landscape and high-level interaction features, trajectory features explicitly exploit the sequence of solutions evaluated by an algorithm during its search process, making them especially suitable for dynamic, run-dependent characterizations of algorithm behavior. Please note that since trajectory features are based on the solutions sampled by the algorithm, we do not include the sampling information as we did for the problem landscape features.

The table shows that most trajectory-based approaches are manually designed, such as Trajectory-ELA, Iterative-ELA, LON, and Probing trajectories, where handcrafted descriptors are extracted from algorithm runs. More recent approaches, such as Opt2Vec, leverage self-supervised learning by embedding trajectories into vector spaces, while methods like ClustOpt apply unsupervised learning to cluster trajectories. This diversity reflects a growing trend to move beyond static problem landscape features toward representations that capture algorithm dynamics.

In terms of inputs, most methods require potential solutions evaluated by an algorithm during its search process, often using population-based algorithms to generate diverse trajectories. An exception is LONs, which are built from multiple runs of a local search algorithm and rely on detecting local optima and their connectivity.

The problem dimensions tested in the literature remain relatively limited. Many studies focus on low to moderate dimensions (e.g., $d \leq 20$), with some extending to $d=30$ or $d=50$ in the case of Opt2Vec and Iterative-ELA. This indicates that trajectory features are still predominantly explored in small-scale benchmarks, and their scalability to higher-dimensional optimization problems remains largely untested.

\begin{table}[!htp]
\centering
\caption{Summary of inputs, training task types and problem dimensions on which the trajectory features have been tested}\label{tab:sum_trajectory}
\scriptsize
\begin{tabular}{p{1.5cm}p{1.5cm}p{5cm}p{5cm}}
\toprule
\textbf{Feature name} & \textbf{Training task type} & \textbf{Inputs} & \textbf{Problem dimensions tested} \\
\midrule
Trajectory-ELA & manually designed & Potential solutions evaluated by an algorithm during its search process & 
- $d \in \{5\}$ ~\cite{adaptive_landcape_analysis,trajectory_ela_performance_regression}

- $d \in \{5, 10\}$~\cite{per_run_algorithm_selection_warmstarting}

- $d \in \{10\}$~\cite{vermetten2023switch}
\\ \midrule
DynamoRep & manually designed & Potential solutions evaluated by a population-based algorithm during its search process & 
- $d \in \{3\}$ ~\cite{10.1145/3583131.3590401}

- $d \in \{3,5,10,20\}$ ~\cite{dynamorep} \\ \midrule
Opt2Vec & unsupervised & Potential solutions evaluated by a population-based algorithm during its search process, Problem class & - $d \in \{10,30,50\}$ ~\cite{opt2vec} \\ \midrule
Iterative-ELA & manually designed & Potential solutions evaluated by a population-based algorithm during its search process & - $d \in \{10,30,50\}$ ~\cite{opt2vec}

- $d \in \{5\}$ ~\cite{dynamorep}
\\ \midrule
LON & manually designed & Multiple runs of a local search algorithm & 
- $d \in \{3,5\}$ ~\cite{adair2019local}

- $d \in \{3,5,8,12,20\}$ ~\cite{mitchell2023local}
\\ \midrule
Probing trajectories & manually designed & Potential solutions evaluated by a population-based algorithm during its search process & - $d \in \{10\}$ ~\cite{renau2024utility} \\ \midrule
ClustOpt & unsupervised & Potential solutions evaluated by a population-based algorithm during its search process & - $d \in \{2,5,10\}$ ~\cite{cenikj2025clustopt}
\\ \midrule

\end{tabular}

\end{table}

\section{Machine Learning Studies that Utilized Meta-features} 
Table~\ref{tab:review} provides an overview of the research focused on problem, algorithm, and trajectory-based features and their applications in problem classification, algorithm selection, performance prediction, and complementarity of benchmark suites. The table is structured into three horizontal parts, one for each of these categories. Each row represents a specific class of features. The columns are divided into three primary aspects. The first aspect identifies the learning tasks in which the features are evaluated, including problem classification, algorithm selection, performance prediction, and visualization/complementarity analysis. The second and third aspects concern the collection of problem instances used in the research, which may originate from established benchmark suites such as BBOB, CEC, or Nevergrad, or be generated by problem instance generators.

We omit a detailed description of each study presented in the table, as most have been previously discussed, and instead provide a single example to illustrate how the table can be interpreted. For instance, the study~\cite{per_run_algorithm_selection_warmstarting} explores per-run algorithm selection using trajectory-ELA features and algorithm features derived from internal algorithm parameters. Accordingly, in the first part of the table, this study is listed in the cells (Trajectory-ELA, Algorithm Selection) and (Internal Algorithm Parameters, Algorithm Selection). For additional context, the same reference~\cite{per_run_algorithm_selection_warmstarting} appears in the second part of the table, corresponding to the BBOB and Nevergrad benchmark suites. These are the benchmark suites where the learning tasks from the first part of the table are tested.


The table results indicate that the majority of studies have utilized ELA problem landscape features with BBOB benchmark problem instances. This prevalence is expected, as developing an ML model for problem classification, algorithm performance prediction, or algorithm selection benefits from having multiple problem instances per problem class that vary by shifting or scaling. This variation allows for training the ML model on one set of problem instances and testing it on another set that differs only in shift or scale. However, using other benchmark suites, such as the CEC benchmark suites, remains challenging. These suites require the development of zero-shot ML models~\cite{xian2017zero} because each problem class is represented by only a single instance.


Recent trends indicate the emergence of novel problem landscape features, such as TLA, DoE2Vec, and TransOpt, which have demonstrated predictive accuracy comparable to ELA features in problem classification tasks. Additionally, there is a growing interest in trajectory-based features, which capture information about the interaction between an algorithm instance and a problem instance.

A common limitation across all studies is the restricted generalization of the developed ML models to other benchmark suites such as CEC and Nevergrad~\cite{nevergrad}, or to instances generated by problem generators. This is evident from the predominant use of the BBOB benchmark suite for learning and evaluation. To address this limitation in the future, it is essential to incorporate problem instances from diverse benchmark suites and various problem generators (which remain underexplored) into a representative learning set to enhance generalization. Additionally, as the landscape of possible problem instances continually evolves with new problem generators, future work should focus on developing ML models within continual learning frameworks~\cite{aljundi2019task,van2019three,wang2023comprehensive}.

\afterpage{%
    \clearpage
    \thispagestyle{empty}
    \begin{landscape}
        \centering 
\begin{table}
\Large
\centering

\caption{\label{tab:review} Summary of works using problem, algorithm, or trajectory-based features and their applications in the domain of algorithm selection for continuous single-objective optimization. The rows are grouped into problem landscape features, algorithm features, and trajectory-based features. The columns are split into three parts; the first part indicates the \textit{learning scenario} where some specific features are evaluated (problem classification, algorithm selection, performance prediction, or visualization/complementarity analysis), the second and third parts correspond to the set of \textit{problem instances} used in the study, with the first group comprising established benchmark suites (BBOB, CEC, Nevergrad) and the other problem collections obtained from instance generators. }
\resizebox{1.4\textwidth}{!}{
\begin{tabular}{|p{7cm}|p{5cm}p{5cm}p{5cm}p{4cm}|p{10cm}p{4cm}p{3cm}|p{2cm}p{2cm}p{4cm}p{3cm}p{3cm}|}
\toprule
& \multicolumn{4}{c}{\textbf{Learning tasks}} & \multicolumn{3}{|c}{\textbf{Benchmark suites}} & \multicolumn{5}{|c|}{\textbf{Problem generators}} \\
\midrule
\multirow{2}{*}{Features} & \multirow{2}{*}{Problem classification} & \multirow{2}{*}{Algorithm selection} & \multirow{2}{*}{Performance prediction} & \multirow{2}{*}{\parbox{4cm}{\centering Visualization /\\ Complementarity}} & \multirow{2}{*}{BBOB} & \multirow{2}{*}{CEC} & \multirow{2}{*}{Nevergrad} & \multirow{2}{*}{ISA} & \multirow{2}{*}{GP} & \multirow{2}{*}{TR} & \multirow{2}{*}{Affine} & \multirow{2}{*}{GKLS} \\
& & & & & & & & & & & & \\
\multicolumn{13}{l}{\cellcolor{lightgray}\textbf{Problem landscape features}} \\
\multirow{4}{*}{ELA} & \multirow{4}{*}{\cite{eftimov2020linear,ela_cell_mapping,ela_ic,tla,renau2020exploratory}}&\multirow{4}{*}
{\parbox{5cm}{\cite{as_survey,10.1145/3583133.3590697,lacroix2019,skvorc2022transfer,jankovic2020landscape,tanabe2022benchmarking, gasper_affine_generalization, gina_generalization_study, gina_hit_the_wall, skvorc2022transfer, gasper_pitfalls}}} & \multirow{4}{*}{\parbox{5cm}{\cite{anak_ela_feature_importance,lacroix2019,10.1145/3583133.3590617,anan_ela_feature_importance_2,anan_ela_feature_importance,risto_ela_feature_importance,nikolikj2023rf+,10254146,10.1145/3583131.3590424,jankovic2020landscape,munoz2012meta,MABBOBteloversion,vanstein2024explainable}}} & \multirow{4}{*}{\parbox{4cm}{\cite{affine_problems,eftimov2020linear,kudela2023computational,lang2021exploratory,long2022learning,long2023challenges} \\ \cite{munoz2020generating,urban_complementary_analysis,MABBOBteloversion,urban_ela_not_invariant,urban_ela_not_invariant_sampling,10.1145/3583131.3590424}

}}& \multirow{4}{*}{\parbox{10cm}{\cite{affine_problems,eftimov2020linear,ela_cell_mapping,as_survey,10.1145/3583133.3590697,jankovic2020landscape,kudela2023computational,lacroix2019,lang2021exploratory,long2022learning,long2023challenges,anak_ela_feature_importance,munoz2012meta,ela_ic,10.1145/3583133.3590617,anan_ela_feature_importance_2,anan_ela_feature_importance,renau2020exploratory,10.1145/3583131.3590424,munoz2020generating,MABBOBteloversion,urban_ela_not_invariant,urban_ela_not_invariant_sampling,risto_ela_feature_importance,urban_complementary_analysis,gina_generalization_study,vanstein2024explainable,tanabe2022benchmarking, skvorc2022transfer, gasper_affine_generalization, gasper_pitfalls}}}&\multirow{4}{*}{\parbox{4cm}{\cite{munoz2012meta,kudela2023computational,lang2021exploratory,10.1145/3583133.3590617,nikolikj2023rf+,10254146} \\ \cite{urban_ela_not_invariant_sampling,urban_ela_not_invariant}}} &\multirow{4}{*}{} &\multirow{4}{*}{ \cite{munoz2020generating}}& \multirow{4}{*}{\cite{long2023challenges}}& \multirow{4}{*}{\cite{long2022learning,urban_complementary_analysis,skvorc2022transfer,gina_generalization_study}}&\multirow{4}{*}{\parbox{3cm} {\cite{affine_problems,MABBOBteloversion,gina_generalization_study, gasper_affine_generalization, gina_hit_the_wall} }}& \multirow{4}{*}{\cite{kudela2023computational}}\\
& & & & & & & & & & & & \\
& & & & & & & & & & & & \\
& & & & & & & & & & & & \\
TLA & \cite{tla,tinytla}  & \cite{tinytla, gina_hit_the_wall} & & & \cite{tla} & & & & & & \cite{gina_hit_the_wall} & \\
Fitness Map + CNNs & \cite{seiler2022collection} &  & & & \cite{seiler2022collection}& & & & & & & \\
Point Cloud Transformer & \cite{seiler2022collection} & & & & \cite{seiler2022collection} & & & & & & & \\
DoE2Vec & \cite{doe2vec} & \cite{gina_hit_the_wall} & & & \cite{doe2vec} & & & & & \cite{doe2vec} & \cite{gina_hit_the_wall} & \\
TransOpt & \cite{cenikj2023transopt} & \cite{gina_generalization_study}  &  & \cite{gina_generalization_study} & \cite{cenikj2023transopt, gina_generalization_study} & & & & &  & \cite{gina_generalization_study} & \\
Deep-ELA & \cite{seiler2024deep} & \cite{seiler2024deep, gina_hit_the_wall, deep_classical_LION, seiler2024learned}  &  & & \cite{seiler2024deep, deep_classical_LION}  & & & & &  & \cite{gina_hit_the_wall,seiler2024learned} & \\
Random Filter Mappings & \cite{random_filters_gasper}  & \cite{random_filters_gasper}  & &  & \cite{random_filters_gasper} & & & & & & & \\
\multicolumn{13}{l}{\cellcolor{lightgray}\textbf{Algorithm features}} \\
Source Code & & \cite{algorithm_code_features} & & & & & & & & & & \\
\multicolumn{13}{l}{\cellcolor{lightgray}\textbf{High-level problem-algorithm interaction features}} \\
Fitness Map + CNNs &  & \cite{prager2021towards} & & & \cite{prager2021towards}& & & & & & & \\
TransOptAS &  & \cite{transoptas,gina_hit_the_wall, seiler2024learned} &  & &  &  & & & & \cite{transoptas} & \cite{gina_hit_the_wall, seiler2024learned} & \\
Performance2Vec & \cite{performance2vec} & & & \cite{performance2vec} & \cite{performance2vec} & & & & & & & \\
Explainable Prediction Models & & & \cite{10.1145/3583133.3590617,anan_ela_feature_importance_2} & &\cite{10.1145/3583133.3590617,anan_ela_feature_importance_2} & & & & & & & \\
Internal Algorithm Parameters & \cite{algorithm_features_time_series_cmaes} & \cite{per_run_algorithm_selection_warmstarting} & \cite{algorithm_features_time_series_cmaes} & & \cite{algorithm_features_time_series_cmaes} & & & & & & & \\
KG embeddings & & & \cite{anak_kg_performance_prediction} & & \cite{anak_kg_performance_prediction} & & & & & & & \\
GNN embeddings & & & \cite{kostovska2025geometric} & & \cite{kostovska2025geometric} & & & & & & & \\
fANOVA & & &  & \cite{nikolikj2024quantifying}& \cite{nikolikj2024quantifying} & & & & & & & \\
fANOVA & & & \cite{vanstein2024explainable} & & \cite{vanstein2024explainable}  & & & & & & & \\
\multicolumn{13}{l}{\cellcolor{lightgray}\textbf{Trajectory-based features}} \\
Trajectory-ELA & \cite{adaptive_landcape_analysis} & \cite{trajectory_ela_performance_regression,per_run_algorithm_selection_warmstarting,vermetten2023switch,jankovic2022trajectory} & & & \cite{adaptive_landcape_analysis,trajectory_ela_performance_regression,per_run_algorithm_selection_warmstarting,vermetten2023switch,jankovic2022trajectory} & & \cite{per_run_algorithm_selection_warmstarting} & & & & & \\
DynamoRep & \cite{10.1145/3583131.3590401, dynamorep} & \cite{dynamorep} & & & \cite{10.1145/3583131.3590401, dynamorep} & & & & & & & \\
Opt2Vec & \cite{korovsec4316939opt2vec} & & & & & \cite{korovsec4316939opt2vec} & & & & & & \\
Iterative-ELA & \cite{korovsec4316939opt2vec, dynamorep} & \cite{dynamorep} & & &\cite{dynamorep}& & & & & & & \\
LON & & & & \cite{adair2019local,mitchell2023local} & \cite{adair2019local,mitchell2023local} & & & & & & & \\
Probing trajectories & & \cite{renau2024utility}& &  & \cite{renau2024utility} & & & & & & & \\
ClustOpt & & & & \cite{cenikj2025clustopt} & \cite{cenikj2025clustopt} & & & & & & & \\
\bottomrule
\end{tabular}
}
\end{table}

\end{landscape}
    \clearpage
}

\section{Discussion and Open challenges} 

In this section, we provide a more detailed discussion of challenges in representation learning of problem, algorithm, and trajectory features for single-objective black-box optimization. 

\subsection{Problem Landscape Features}

The calculation of ELA features involves the application of a set of statistical functions to candidate solutions that are artificially sampled from the decision space of the optimization problem instance. Despite their prevalence, the calculation of ELA features can be computationally demanding for high-dimensional problems, and they have been shown to be sensitive to variations in the sample size and sampling method~\cite{renau2020exploratory, urban_ela_not_invariant_sampling}, as well as not being invariant to transformations such as scaling and shifting of the optimization problem~\cite{urban_ela_not_invariant,urban_ela_not_invariant_sampling}. The TLA and TinyTLA features have the desired property of being invariant to these transformations and have been demonstrated to have good predictive performance for the classification of optimization problems. While an initial study of the predictive performance of the TinyTLA features for the algorithm selection task has been conducted on the BBOB benchmark, a more comprehensive evaluation can be done, involving different algorithm portfolios and benchmark sets.

With the rapidly increasing number of studies that propose low-level landscape features based on deep learning, they come with the limitation of not allowing interpretation of why a decision is made. There is also no clear evidence of which features are the best for different learning tasks or if they are in favor of some algorithm classes.
With the rapidly growing body of deep learning–based low-level landscape features, a persistent limitation is their limited interpretability: they rarely explain why a particular decision is made. Moreover, there is still no clear evidence identifying which feature sets are best for specific learning tasks, or whether certain features systematically favor particular algorithm classes.
Several works have directly compared problem landscape features. 
A comparison between TransOpt (trained for problem classification) and classical ELA features is reported in~\cite{gina_generalization_study}. In this work, the generalization of algorithm selection models across four benchmark suites is being evaluated with different feature groups. The results produce mixed outcomes of the superiority of one feature group over another, showing that the quality of predictions depends heavily on the benchmarks used for training and testing, and the distribution of algorithm performance.
Another comparison of the DeepELA and ELA features appears in~\cite{deep_classical_LION}, where the features are evaluated for the algorithm selection task in a fixed-target setting on the BBOB benchmark. In this case, a benefit of the joint use of ELA and DeepELA features is observed. However, when the DeepELA, ELA and TransOptAS features are evaluated in a fixed-budget setting on the affine combinations of the BBOB problems~\cite{MABBOBteloversion}, the ELA features are shown to be superior over the transformer-based ones~\cite{seiler2024learned}.
The most comprehensive comparative study including the ELA, TransOptAS, DeepELA, Doe2Vec and TinyTLA features is provided in~\cite{gina_hit_the_wall}. In this study, features are compared for the algorithm selection task on four different algorithm portfolios using the affine combinations of the BBOB benchmark problems~\cite{MABBOBteloversion}. The algorithm selection models are evaluated in a fixed-budget setting, on evaluation settings of increasing difficulty. The general outcome is that the ELA features remain superior to newer problem landscape features in easy evaluation settings, however, none of the features outperform a baseline model in the difficult evaluation settings. This indicates a need to rethink the standard algorithm selection setup using problem landscape features, as well as a need for improved benchmarks suites for evaluation.
Overall, we can observe that various aspects of the experimental setup (problem benchmarks, method for splitting data into training and test sets, algorithms included in algorithm portfolio, sample size used for feature calculation, metrics used for capturing algorithm performance) used for comparing problem landscape features for the algorithm selection task can substantially influence on the results of the comparison. Therefore, standardization of the experimental setup is needed to ensure that comparisons of problem landscape features for the algorithm selection task are consistent, reliable, and reproducible.

\subsection{Algorithm Features}

Algorithm features are relatively underexplored and low-resourced compared to other feature types. A promising direction is to represent algorithms as modular frameworks, similar to modCMA and modDE, by structuring them into components that reflect their operators. Such insights could then be utilized as meta-features, complementing problem-, high-level problem-algorithm interaction, or trajectory-based features, and applied to various machine learning tasks. Proper algorithm features would also allow us to go beyond the classical taxonomy-style representation of algorithms~\cite{stork2020new} to more complex descriptions that allow to cluster algorithms according to different criteria.  

\subsection{High-Level Problem-Algorithm Interaction Features}
For offline automated algorithm selection and configuration methods, these features are crucial. However, most of these features require the prior computation of ELA features and/or algorithm performance, which is itself computationally demanding. This is the case for the GNN and KG embeddings as well as the Explainable Prediction Models, requiring both ELA features and algorithm performance. On the other hand, the TransOptAS and Fitness Map + CNNs features require algorithm performance for a sufficiently large set of functions to train a neural network, while the fANOVA features require algorithm performance for a sufficiently large coverage of the parameter space of the analyzed algorithm.

Future directions should focus on model-agnostic approaches to characterize the interactions between problems and algorithms. Furthermore, since supervised machine learning is typically applied in such learning processes, the quality of the data used for training is of great importance.


\subsection{Trajectory-based Features}
The process of automated algorithm selection and configuration could potentially benefit from features that take into account the interactions between the problem and the algorithm. Moreover, utilizing samples from the optimization algorithm's trajectory incurs no additional computational expenses, since there is no need to evaluate the objective function of the optimization problem prior to running the algorithm. A potential direction for future research would involve the generation of new features tailored specifically for online use, which would be computationally inexpensive enough to be calculated during algorithm execution. Despite the DynamoRep (statistical) and Opt2Vec (based on autoencoder) features, features from the trajectory of an optimization algorithm instance could be extracted using expert knowledge or deep learning methods such as Long Short-Term Memory Networks (LSTM), Convolutional Neural Networks (CNN), or Transformers.

While trajectory-based features provide rich information about problem–algorithm interactions, their computational costs can vary, which is crucial for assessing online feasibility and the type of tasks they are suited for. Features derived from internal algorithm parameters, probing trajectories or simple population statistics as in DynamoRep incur negligible per-iteration overhead and are well suited for online use. Iterative-based ELA are more demanding since they perform the calculation of the ELA features on every iteration of the algorithm, and Local Optima Networks or Search Trajectory Networks are even heavier, relying on repeated local search or basin-hopping steps, which limits them to offline analysis. ClustOpt features perform clustering only once across all iterations, resulting in a moderate one-time cost, making it practical for online monitoring of algorithm execution or post-hoc analysis of multiple algorithm runs. To enable early-run selection or warm-start switching, lightweight proxies such as probing trajectories features, DynamoRep or trajectory ELA features can be explored, although their predictive power is yet to be evaluated for this task.

Trajectory features also rely on the quality and diversity of samples generated by the underlying population-based algorithm. Because the statistical reliability of the learned features depends strongly on this data, the lack of standardized benchmarking introduces ambiguity in the reported results.

It is important to note that trajectory-based features are recently developed and comprehensive comparative analyses are yet to be performed. With this in mind, it is difficult to provide a definite recommendation of the use of one feature group over another.

\subsection{Challenges for Using the Meta-features for Machine Learning}

It seems widely accepted in the evolutionary computation community that random forest models~\cite{jankovic2021impact} in combination with ELA problem landscape features provide acceptable results. However, we also observe a trend that recent advances in ML modeling are taken up by the community only with some delay or are neglected after some nonsatisfactory results on the BBOB benchmark suite. We are concerned that the good results achieved on this comparatively small problem collection may paint a too optimistic picture of the capability of ELA and simple random forest models. This concern seems to be confirmed by recent works pointing out the rather dissatisfying generalization ability of current algorithm selection models based on ELA features~\cite{skvorc2022transfer, gasper_affine_generalization, gina_generalization_study, tanabe2022benchmarking, gina_hit_the_wall}. This has been demonstrated in a scenario where the algorithm selection model trained on one benchmark and evaluated on another one does not outperform a simple baseline model~\cite{gina_generalization_study, skvorc2022transfer}. Additionally, it has been shown that algorithm selection models perform well when similar problem instances are present in the training and testing set, but fail when presented with unseen problem instances~\cite{gasper_affine_generalization}. Finally, the leave-one-instance-out evaluation strategy, which is the most commonly used to evaluate algorithm selection models on the BBOB benchmark suite, has been criticized as it can produce misleading results, where meta-models achieve high performance due to spurious correlations between features and the target, rather than having genuine predictive capability~\cite{gasper_pitfalls}. The use of scale-sensitive metrics to capture algorithm performance has also raised concerns for causing a false indication of improvements over baseline models used in algorithm selection~\cite{gasper_pitfalls}.

To improve the generalizability of algorithm selection models, several avenues can be explored: including newly developed features, including larger, more diverse training data, and considering contemporary ML training strategies.
 
To get a fair comparison of feature performance, comprehensive studies are required which evaluate different feature sets under a fixed experimental setup, using consistent datasets, splits, and evaluation protocols to enable meaningful statistical testing and reliable conclusions. Such studies should further include diverse problem benchmarks beyond BBOB, report variability and deviations in performance, establish proper baselines for comparison, and analyze the influence of sample sizes and sampling techniques. Current research relies disproportionately on a narrow set of benchmark datasets (primarily BBOB), and comparisons conducted within the same benchmark often produce dependent performance values, making many of the reported findings unsuitable for rigorous statistical testing. The need to rethink existing benchmark suites has been raised across the broader field~\cite{rethinking_benchmarking}.

Finally, we believe it is important to critically observe advances in ML utilizing other models based on deep learning architectures that can be evaluated in transfer learning and continual learning scenarios. Continual learning~\cite{van2019three,aljundi2019task,wang2023comprehensive} is a ML paradigm that allows models to develop themselves adaptively by learning incrementally from dynamic data distributions and selectively adapting representations to obtain good generalizability within and between tasks. This adaptation can lead to a range of benefits such as improved robustness by handling both simple and complex data instances, reducing overfitting to specific data instances, resulting in better generalization to unseen data. By using continual learning approaches, one hopes to identify problem instances that challenge the performance predictor assumptions and explore the uncharted regions of the problem space, ultimately leading to algorithm selection models that not only perform well on the training problem instances, but also generalize effectively to unseen problem instances.

\subsection{Practical recommendations.}
For practitioners, the choice of feature family should primarily depend on what needs to be modeled (the problem, the algorithm, or their interaction), what data is available, the evaluation budget, the need for invariance/robustness, and whether the goal is offline modeling or online decision-making. If a well-tested and broadly applicable baseline for offline tasks (e.g., instance classification or algorithm selection) is needed and an initial sampling phase is affordable, ELA remains a sensible first choice. Deep learned landscape features can also be used when interpretability is not a primary requirement. When applying problem landscape features to the algorithm selection task, one should not expect a good generalization to unseen problems, i.e., the model will only perform well if training and testing problems are nearly identical, and the model may need to be retrained on the particular types of problems it will be applied on.
When aspects of algorithm performance (rather than only landscape structure) need to be incorporated into the feature representation, high-level problem-algorithm interaction features can be effective; however, some of these are computationally expensive to compute, as detailed in previous sections. Finally, when extra pre-evaluations should be avoided and/or information computed during the run is required (e.g., early-run selection, monitoring, or switching), trajectory-based features are a natural option: DynamoRep and probing-trajectory features can be computed online with minimal overhead, whereas heavier iterative-ELA or Local Optima Network variants are better reserved for offline analysis when computational cost is less constrained. In this context, approaches such as ClustOpt and Local Optima Networks are particularly useful for visualizing algorithm behaviour and relating search dynamics to structural properties of the problem.



\bibliography{references}

\end{document}